\documentclass[journal]{IEEEtran}
\usepackage{amsmath,amsfonts}
\usepackage{algorithmic}
\usepackage{algorithm}
\usepackage{array}
\usepackage[caption=false,font=normalsize,labelfont=sf,textfont=sf]{subfig}
\usepackage{textcomp}
\usepackage{stfloats}
\usepackage{url}
\usepackage{verbatim}
\usepackage{graphicx}
\usepackage{cite}
\usepackage{multirow}
\usepackage[pagebackref,breaklinks,colorlinks,allcolors=blue]{hyperref}
\usepackage{booktabs} % 建议使用以提升表格质量
\usepackage{makecell}
\usepackage{pifont}
\begin{document}

\title{SDF-Net: Structure-Aware Disentangled Feature Learning for Optical--SAR Ship Re-Identification}

\author{Furui~Chen, Han~Wang, Yuhan~Sun, Jianing~You, Yixuan~Lv, Zhuang~Zhou, Hong~Tan, and~Shengyang~Li
    \thanks{This work was supported by the Strategic Priority Research Program of the Chinese Academy of Sciences under Grant XDA0360303. \textit{(Corresponding author: Shengyang Li.)}}
    \thanks{Furui Chen, Han Wang, Yuhan Sun, Jianing You, Hong Tan, and Shengyang Li are with the Technology and Engineering Center for Space Utilization, Chinese Academy of Sciences, Beijing 100094, China, also with the Key Laboratory of Space Utilization, Chinese Academy of Sciences, Beijing 100094, China, and also with the University of Chinese Academy of Sciences, Beijing 100049, China (e-mail: chenfurui24@mails.ucas.ac.cn; wanghan221@mails.ucas.ac.cn; sunyuhan21@mails.ucas.ac.cn; youjianing24@csu.ac.cn; tanhong@csu.ac.cn; shyli@csu.ac.cn).}
    \thanks{Yixuan Lv is with the Technology and Engineering Center for Space Utilization, Chinese Academy of Sciences, Beijing 100094, China, and also with the Key Laboratory of Space Utilization, Chinese Academy of Sciences, Beijing 100094, China, and also with the School of Software, Beihang University, Beijing 100191, China(e-mail: lvyixuan@csu.ac.cn).}
    \thanks{Zhuang Zhou is with the Technology and Engineering Center for Space Utilization, Chinese Academy of Sciences, Beijing 100094, China, and also with the Key Laboratory of Space Utilization, Chinese Academy of Sciences, Beijing 100094, China (e-mail: zhouzhuang@csu.ac.cn).}
}

\markboth{EEE Journal of Selected Topics in Applied Earth Observations and Remote Sensing}%
{Chen \MakeLowercase{\textit{et al.}}: SDF-Net: Structure-Aware Disentangled Feature Learning for Optical--SAR Ship Re-identification}

% \IEEEpubid{0000--0000/00\$00.00\copyright2021 IEEE}

\maketitle

\begin{abstract}
Cross-modal ship re-identification (ReID) between optical and synthetic aperture radar (SAR) imagery is fundamentally challenged by the severe radiometric discrepancy between passive optical imaging and coherent active radar sensing. While existing approaches primarily rely on statistical distribution alignment or semantic matching, they often overlook a critical physical prior: ships are rigid objects whose geometric structures remain stable across sensing modalities, whereas texture appearance is highly modality-dependent. In this work, we propose SDF-Net, a Structure-Aware Disentangled Feature Learning Network that systematically incorporates geometric consistency into optical--SAR ship ReID. Built upon a ViT backbone, SDF-Net introduces a structure consistency constraint that extracts scale-invariant gradient energy statistics from intermediate layers to robustly anchor representations against radiometric variations. At the terminal stage, SDF-Net disentangles the learned representations into modality-invariant identity features and modality-specific characteristics. These decoupled cues are then integrated through a parameter-free additive residual fusion, effectively enhancing discriminative power. Extensive experiments on the HOSS-ReID dataset demonstrate that SDF-Net consistently outperforms existing state-of-the-art methods. The code and trained models are publicly available at \url{https://github.com/cfrfree/SDF-Net}.
\end{abstract}


\section{Introduction}
\IEEEPARstart{O}{ptical} and Synthetic Aperture Radar (SAR) sensors play complementary roles in maritime surveillance. Optical imagery provides rich visual details under favorable illumination, whereas SAR, as an active sensing modality, enables all-weather and day-night observation by measuring microwave backscatter. Integrating these heterogeneous sources is therefore critical for continuous ship monitoring and long-term target tracking \cite{9416740}. As a fundamental component of this integration, cross-modal ship re-identification (ReID) aims to associate ship identities across optical and SAR imagery \cite{schmitt2019sen12ms, Wang_2025_ICCV}.

\begin{figure}[t]
    \centering
    \includegraphics[width=\linewidth]{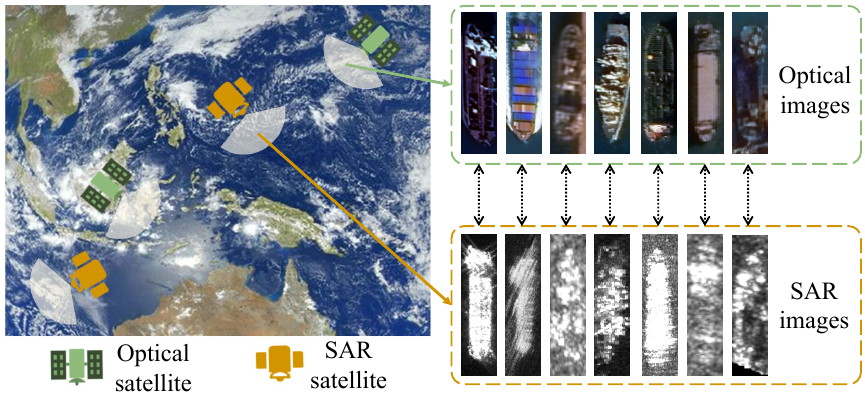}
    \caption{Illustration of the optical--SAR ship re-identification task. The framework aims to match the same ship identity across vastly different optical and SAR sensor modalities.}
    \label{fig:task_illustration}
\end{figure}

Optical--SAR ship ReID must be distinguished from the more extensively studied visible--infrared person ReID (VI-ReID). In VI-ReID, both sensors operate in the passive electro-optical regime---visible at $\sim$0.4--0.7~$\mu$m and thermal infrared at $\sim$8--14~$\mu$m---and appearance attributes such as clothing color and texture exhibit partial consistency across modalities~\cite{zheng2022visible}. VI-ReID research therefore emphasizes deformable pose variation and part-level correspondence. Optical--SAR ship ReID departs from this framework in three fundamental ways.

The first is the sensing physics itself. SAR is an active microwave sensor operating at centimeter-scale wavelengths such as X-band and C-band, measuring coherent backscatter governed by surface roughness, dielectric properties, and corner-reflector geometry. Optical sensors passively collect solar reflectance at sub-micron wavelengths governed by material absorption spectra and illumination conditions. This is not a difference of degree but of physical kind: the two modalities measure fundamentally different quantities. The second difference lies in modality-specific artifacts. SAR imagery is corrupted by multiplicative speckle noise~\cite{zhang2020ls}, discrete high-intensity corner reflector responses from metallic superstructures, and geometric distortions---layover, foreshortening, and shadowing---induced by the side-looking range-Doppler imaging geometry~\cite{schmitt2019sen12ms}. Optical imagery suffers from cloud occlusion, sunglint on water surfaces, and illumination variation due to time-of-day and season. These artifacts are unique to each modality and cannot be aligned through statistical matching alone. The third difference concerns object deformation. Unlike pedestrians with deformable poses, ships are rigid structures whose macroscopic geometry---hull contour, aspect ratio, superstructure layout---remains stable across sensor modalities and viewing conditions, particularly under the near-nadir observation geometry typical of LEO satellite imaging. This physical prior is the foundation of our approach.

These properties call for a solution that goes beyond purely data-driven statistical alignment. A physics-guided approach, one that explicitly anchors cross-modal association on invariant geometric structures while accommodating modality-specific radiometric distortions, is required.

However, optical--SAR ReID remains highly challenging due to the intrinsic physical disparity between the two sensing mechanisms. This massive modality gap induces complex non-linear radiometric distortion (NRD) \cite{8314449}, characterized by fundamentally inconsistent intensity responses across sensors due to the vast wavelength difference between microwave backscattering and visible reflectance. Consequently, traditional feature alignment predicated on metric-based distance becomes mathematically ill-posed, as direct appearance correspondence is corrupted by modality-specific signal fluctuations rather than simple Gaussian noise \cite{merkle2018exploring}. While optical images capture passive reflectance patterns governed by illumination and material properties, SAR imagery is fundamentally dominated by coherent scattering effects and speckle noise. Texture appearance thus exhibits severe modality-specific distortion, a challenge extensively documented in radar-based ship analysis \cite{zhang2020ls}, rendering direct appearance alignment unreliable and often misleading.

Most existing approaches address this challenge from a data-driven perspective, formulating cross-modal ReID as a feature distribution alignment problem \cite{pami21reidsurvey, Liu_2022_CVPR}. Early works focus on learning a shared embedding space to reduce modality gaps, while recent deep models leverage convolutional neural networks or Vision Transformers (ViT) to implicitly align high-level semantic representations \cite{he2021transreid}. Although these methods achieve encouraging performance, they typically treat feature extraction as a black-box process and lack explicit mechanisms to distinguish modality-invariant identity cues from sensor-specific interference such as speckle, sea clutter, or illumination variations.

To further mitigate modality discrepancy, prevailing research has explored generative synthesis and intricate distribution matching strategies \cite{zhu2017unpaired, li2022cross}. Although these approaches reduce statistical divergence, they frequently impose prohibitive computational costs and risk introducing hallucinatory artifacts that obscure identity-critical features \cite{yang2022sar}. Critically, such purely statistical alignment overlooks the physical constraints of maritime targets as rigid bodies. In contrast to pedestrian re-identification—which typically exploits deformable pose alignment or stable anatomical proportions \cite{pami21reidsurvey}—maritime targets exhibit strong intrinsic geometric rigidity. We recognize that SAR imaging inherently introduces projection distortions such as layover and foreshortening under varying incidence angles \cite{schmitt2019sen12ms}. However, while these radar-specific phenomena alter micro-level pixel correspondences, the macro-topological layout, global hull proportions, and superstructure configurations of the ship maintain a high degree of cross-modal consistency, particularly under the overhead, near-vertical observation perspectives typical of satellite remote sensing. Within this near-nadir sensing environment, the macroscopic geometric skeleton provides a robust and distortion-tolerant physical invariant, serving as a definitive anchor for cross-modal representation alignment.

We argue that a more principled solution should directly leverage geometric structure as the common denominator between optical and SAR imagery. For ships, attributes such as hull contour, aspect ratio, and spatial layout are largely invariant across modalities, whereas texture patterns are inherently sensor-dependent. Therefore, enforcing strict consistency on geometric structure while allowing flexibility in modality-specific appearance is crucial for reliable cross-modal ReID. From a representation learning perspective, such geometric information is neither best captured at the raw pixel level nor at highly abstract semantic layers. Instead, it is preserved in intermediate network representations that retain spatial organization while being sufficiently abstracted from low-level noise. This observation motivates us to formally model and constrain structural consistency at intermediate feature layers, rather than relying on implicit alignment at the output level.

Based on these insights, we propose SDF-Net, a Structure-Aware Disentangled Feature Learning Network for optical--SAR ship re-identification. Built upon a ViT backbone, SDF-Net introduces a Structure Consistency Constraint to enforce cross-modal geometric alignment at intermediate stages.

At the terminal stage, SDF-Net further decouples the learned representations into modality-invariant shared features and modality-specific features \cite{zheng2019joint}. To effectively integrate these complementary components, we employ an additive fusion strategy, where modality-specific features act as a residual refinement to the shared identity representation. This simple yet effective design avoids feature redundancy and preserves discriminative identity information without introducing additional computational overhead.

In this paper, the term ``physics-guided'' refers to the direct integration of the physical properties of maritime targets and sensor mechanisms into the network design. First, exploiting the physical prior that ships are rigid bodies, we introduce a Structure Consistency Constraint to anchor the cross-modal alignment on the invariant geometric hull rather than highly variable textures. Second, to address the distinct physical imaging mechanisms---specifically, the high-intensity dynamic range of SAR coherent scattering versus the narrow-band diffuse reflectance of optical sensors---we apply instance normalization to the gradient energy. This mathematically standardizes the disparate amplitude responses into a modality-agnostic structural descriptor, forming a cohesive physics-informed representation learning paradigm.

The main contributions of this work are summarized as follows:
\begin{itemize}
    \item We propose SDF-Net, a physics-guided representation learning framework for optical--SAR ship re-identification, which moves beyond implicit statistical matching by firmly anchoring cross-modal association on invariant geometric structures.
    \item We introduce a scale-invariant structure consistency constraint based on normalized gradient energy statistics from intermediate Transformer layers, enabling highly robust alignment against severe radiometric distortions.
    \item We design a disentangled feature learning and additive fusion strategy that seamlessly integrates modality-specific residual information into shared identity representations in a completely parameter-free manner.
    \item Extensive experiments on the HOSS-ReID dataset \cite{Wang_2025_ICCV} demonstrate that the proposed method achieves state-of-the-art performance, thoroughly validating the efficacy of physics-guided disentanglement for maritime target association.
\end{itemize}
\section{Related Work}

In this section, we review research efforts closely related to the proposed SDF-Net, organized into three research streams: cross-modal re-identification, disentangled and structure-aware representation learning, and optical--SAR ship analysis.

\subsection{Cross-Modal Re-Identification}

Cross-modal re-identification (ReID) has undergone extensive development within the framework of visible--infrared person ReID (VI-ReID), which focuses on mitigating the distribution gaps between heterogeneous sensors through the acquisition of modality-robust representations \cite{pami21reidsurvey}. While early methodologies primarily treated modality discrepancy as noise to be suppressed within a shared embedding space, more recent investigations suggest that modality-specific information contains vital discriminative cues that can be selectively preserved to refine identity representations under appropriate constraints \cite{Ren_2024_CVPR}.

To address spatial misalignment and local correspondence ambiguity, semantic alignment and affinity reasoning mechanisms have been employed to aggregate consistent local regions across disparate manifolds \cite{Fang_2023_ICCV}. Furthermore, the emergence of multimodal contrastive paradigms has facilitated the alignment of cross-modal positive pairs by pulling them together within a unified hypersphere manifold \cite{zhang2021attend}. In parallel, structural priors have been integrated into Transformer-based architectures to safeguard spatial integrity across modalities \cite{9725265}.

Despite the maturity of VI-ReID, its design principles do not transfer directly to optical--SAR ship ReID. We identify three categorical mismatches that limit the applicability of existing methods to this domain.

The first is the sensing physics. VI-ReID operates entirely within the passive electro-optical regime, covering visible and thermal infrared wavelengths that measure emitted or reflected radiation in the $\sim$0.4--14~$\mu$m range; the modality gap is primarily a spectral emissivity difference. Optical--SAR ReID spans fundamentally different physical mechanisms: active coherent microwave backscatter at $\sim$3--5.6~cm SAR wavelengths, sensitive to surface roughness and dielectric constant, versus passive solar reflectance at $\sim$0.4--0.9~$\mu$m optical wavelengths, sensitive to material albedo and illumination. The resulting non-linear radiometric distortion (NRD)~\cite{8314449} is orders of magnitude more severe, and the statistical assumptions underlying VI-ReID distribution alignment such as Gaussian noise models, break down under SAR's multiplicative speckle and discrete corner-reflector responses~\cite{zhang2020ls}. The second concerns object deformation. VI-ReID methods invest substantial capacity in modeling human pose variation and part-level articulation~\cite{zheng2022visible}. Ships, as rigid maritime structures, exhibit negligible intra-identity geometric deformation under operational satellite viewing geometries. The challenge is inverted: instead of modeling deformation, the network must exploit geometric stability as an anchor while tolerating extreme radiometric variation, a capability that VI-ReID architectures are not designed to provide. The third concerns the role of modality-specific information. In VI-ReID, modality-specific features such as thermal-specific texture, are routinely discarded as nuisance variance~\cite{choi2020hi}. In optical--SAR ReID, these features carry physically meaningful information: SAR corner-reflector responses encode superstructure geometry, while optical color and texture encode material properties. Discarding either modality's signal as noise removes identity-relevant physical evidence. These modality-specific cues should instead be preserved and integrated as complementary information, which motivates the additive residual fusion design in SDF-Net.

These mismatches are not merely quantitative, reflecting a larger modality gap, but qualitative: the underlying physics, the target deformation model, and the role of modality-specific information are all categorically different. Existing VI-ReID methods, regardless of their architectural sophistication, fail to address these structural differences because they are optimized for a problem with fundamentally different physical constraints. Our physics-guided approach directly targets the specific properties of optical--SAR ship ReID: geometric anchoring for rigid targets, gradient energy normalization for active--passive intensity disparity, and residual fusion for preserving complementary sensor signatures.

\subsection{Disentangled and Structure-Aware Learning}

Disentangled representation learning seeks to decouple identity-relevant information from modality-dependent variations. In the general re-identification (ReID) domain, generative and factorized models have been employed to separate structural content from appearance attributes \cite{zheng2019joint}. While feature disentanglement has been extensively explored in visible-infrared person ReID (VI-ReID) architectures like Hi-CMD \cite{choi2020hi}, these mature strategies cannot be trivially adapted to optical--SAR ship ReID. VI-ReID fundamentally aims to decouple deformable human poses from modality-specific clothing colors under similar passive imaging mechanisms. In stark contrast, maritime targets are rigid bodies, yet their cross-modal discrepancy stems from severe non-linear radiometric distortions caused by distinct physical imaging mechanisms---active microwave coherent scattering versus passive diffuse reflectance. Consequently, rather than discarding the modality-specific feature $\mathbf{f}_{\text{sp}}$ as mere style noise as commonly practiced in VI-ReID, our approach uniquely treats $\mathbf{f}_{\text{sp}}$ as a physical sensor footprint (e.g., SAR corner reflector responses) and preserves it via an additive residual fusion to complement the rigid geometric skeleton $\mathbf{f}_{\text{sh}}$. For instance, hierarchical cross-modal disentanglement architectures \cite{choi2020hi} have demonstrated substantial efficacy in formally factorizing representations into modality-invariant identity features and modality-specific style codes. However, while shape-erased or pose-invariant representations have proven effective for articulated objects such as humans, these assumptions are less applicable to rigid targets. For ships, geometric structure remains largely invariant across viewpoints and sensing modalities, providing a reliable anchor for cross-modal alignment. Furthermore, the integration of Instance Normalization (IN) has been demonstrated to effectively filter out modality-related ``style'' variations while preserving essential content information \cite{pan2018two}.

Beyond disentanglement and normalization-based strategies, recent studies have highlighted the importance of intermediate feature statistics for cross-domain alignment. In neural style transfer, Gram matrix statistics extracted from intermediate convolutional layers are shown to effectively characterize structural and style information \cite{gatys2016image}. Similarly, feature-level statistical matching has been employed to enforce domain consistency by aligning intermediate representations rather than solely relying on output embeddings \cite{zhang2018unreasonable}.

These findings suggest that intermediate layers preserve spatial organization and structural cues that are partially invariant to low-level appearance variations. Inspired by this line of research, we specifically extract normalized gradient energy statistics from intermediate Transformer layers to construct scale-invariant structural descriptors for cross-modal ship ReID.

Nevertheless, completely discarding modality-specific characteristics may lead to the loss of fine-grained discriminative details. This has prompted recent studies to investigate the integration of explicit geometric cues with appearance features. The reliability of geometric structural properties as modality-invariant descriptors is well-established in classical remote sensing matching tasks, such as the Histogram of Oriented Phase Congruency (HOPC) \cite{isprs-annals-III-1-9-2016}, which leverages local geometric structures to bridge the radiometric gap between optical and SAR imagery. Similarly, Xu \emph{et al.} \cite{10687987} demonstrate that incorporating geometric information can effectively reduce modality discrepancy when appropriately fused with learned representations. Building upon these insights, SDF-Net emphasizes modality-invariant geometric signatures as the core structural anchor while maintaining complementary modality-specific information through an additive residual fusion strategy. This approach achieves a balanced trade-off between robustness to non-linear radiometric distortions and the preservation of discriminative ship identity details.

\subsection{Optical--SAR Image Analysis and Ship ReID}

The development of cross-modal ship re-identification has historically been constrained by the scarcity of standardized benchmarks and large-scale annotated datasets. The recent introduction of the CMShipReID dataset \cite{xu2025cmshipreid} has facilitated ship retrieval across visible, near-infrared, and thermal infrared modalities. However, these sensing modalities are predominantly passive and do not reflect the pronounced modality discrepancy introduced by active microwave imaging. The release of the HOSS-ReID dataset \cite{Wang_2025_ICCV} addresses this limitation by providing a dedicated benchmark for optical--SAR association under diverse scattering conditions and complex maritime environments.

Early efforts in bridging the passive-active sensing gap primarily focused on patch-level matching utilizing pseudo-Siamese networks \cite{hughes2018identifying}. Transitioning from generic patch association to instance-level ship retrieval, existing optical--SAR ReID approaches generally fall into three methodological paradigms.

The first paradigm relies on implicit attention-based alignment mechanisms. Representative methods such as TransOSS \cite{Wang_2025_ICCV} employ Vision Transformer architectures with specialized tokenization strategies to model global contextual dependencies across modalities. In these approaches, cross-modal correspondence is expected to emerge implicitly from self-attention modeling. However, relying solely on unconstrained self-attention renders the network highly susceptible to modality-specific distractors. Without a definitive physical anchor, the attention mechanism is frequently misled by the discrete, extremely high-intensity corner reflectors in SAR imagery, or complex hydrodynamic wakes in optical data, leading to severe alignment failures. The proposed SDF-Net addresses this critical bottleneck by enforcing a structural consistency constraint, establishing a reliable geometric anchor that prevents the self-attention manifold from collapsing into modality-specific radiometric noise.

The second paradigm focuses on statistical or generative alignment strategies. Inspired by cross-domain adaptation and image translation techniques, these methods attempt to reduce cross-modal distribution divergence through feature-level matching \cite{long2015learning} or adversarial image translation frameworks such as CycleGAN \cite{zhu2017unpaired}. While such approaches can alleviate global modality gaps, they primarily operate at the distribution level and may introduce artificial artifacts or overlook physically grounded structural invariants that remain stable across sensing mechanisms.

The third paradigm incorporates geometry- or physics-guided structural priors into representation learning. In classical optical--SAR matching literature, structural descriptors such as HOPC \cite{isprs-annals-III-1-9-2016} have demonstrated the effectiveness of leveraging modality-invariant geometric cues to bridge radiometric disparities. Similarly, remote sensing studies have emphasized that structural characteristics are more reliable than raw intensity patterns when associating optical and SAR imagery \cite{schmitt2019sen12ms}.

Unlike attention-based approaches that implicitly expect geometric correspondence to emerge from global context modeling, or generative methods that attempt to reduce statistical divergence at the distribution level, the proposed SDF-Net encodes structural invariance as a core learning objective. This design grounds cross-modal alignment in physically meaningful geometric primitives rather than relying solely on representation-level similarity. By anchoring identity learning in modality-invariant structural priors and integrating modality-specific cues through residual refinement, the proposed framework establishes a physics-informed representation space. Such a physics-guided, structure-aware formulation remains largely underexplored in optical--SAR ship ReID.

\section{Methodology}

\subsection{Problem Definition and Formulation}
The objective of optical--SAR ship re-identification is to establish a robust associative mapping between heterogeneous sensing manifolds. Formally, we define a training dataset as $\mathcal{D} = \{(I_k, y_k, m_k)\}_{k=1}^{N}$, where $I_k \in \mathbb{R}^{H \times W \times 3}$ denotes the $k$-th input image, $y_k \in \{1,\dots,K\}$ represents the ground-truth identity label from a gallery of $K$ distinct ships, and $m_k \in \{0,1\}$ serves as the modality indicator, with $0$ and $1$ signifying optical and SAR domains, respectively.

Unlike single-modality retrieval, cross-modal ship ReID requires the network to transcend the massive radiometric gap while preserving identity-critical geometric signatures. We aim to learn a nonlinear mapping function $\Phi: I \to \mathbf{f}$ that projects raw pixels into a unified, modality-invariant latent embedding space $\mathbb{R}^{d}$. In this optimized manifold, the learned representations must adhere to a dual constraint: minimizing intra-class variance to facilitate cross-modal identity matching while maximizing inter-class separation to ensure high-fidelity discriminative precision across complex maritime backgrounds.

\subsection{Architectural Overview}

\begin{figure*}[t]
    \centering
    \includegraphics[width=\textwidth]{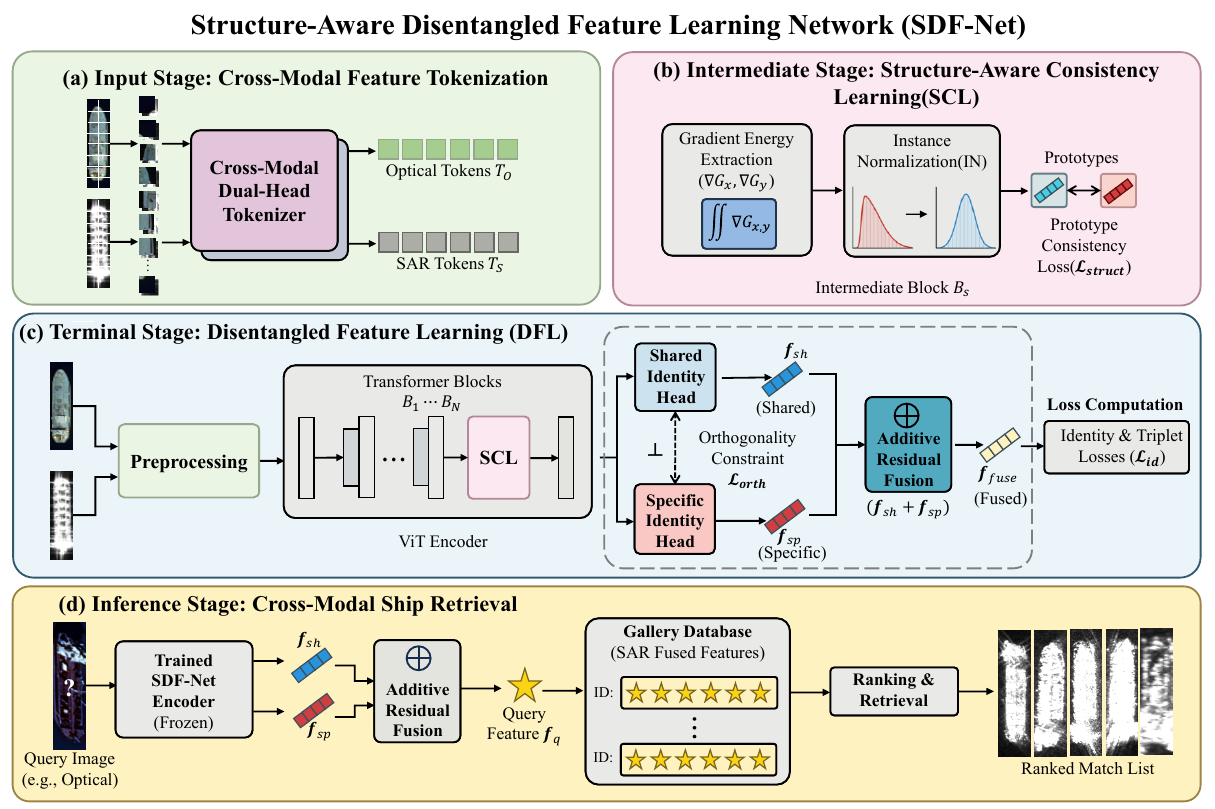}
    \caption{Architectural pipeline of the proposed SDF-Net. The framework builds upon the TransOSS base architecture (cross-modal dual-head tokenizer, MIE, SSE, and ViT-B/16 backbone), initialized with weights pre-trained on optical--SAR image pairs from the TransOSS framework. SDF-Net introduces two novel physics-guided modules: (b) Structure-Aware Consistency Learning (SCL), which extracts and aligns intermediate gradient energy to anchor representations on modality-invariant structural primitives; and (c) Disentangled Feature Learning (DFL), which decouples shared identity features from modality-specific features and fuses them via additive residual refinement. In the ablation study, the baseline disables both SCL and DFL, providing a controlled comparison under the same hyperparameter configuration.}
    \label{fig:framework}
\end{figure*}

As illustrated in Fig. \ref{fig:framework}, the proposed SDF-Net is architected upon a Vision Transformer backbone, chosen for its superior capacity to model the global contextual dependencies essential for capturing the elongated structural properties of maritime targets.

SDF-Net inherits the base architecture of TransOSS~\cite{Wang_2025_ICCV}, including the ViT-B/16 backbone, the cross-modal dual-head tokenizer (separate patch embedding layers for optical and SAR inputs), the Modality Information Embedding (MIE), and the Spatial-Scale Encoding (SSE). These components provide the cross-modal tokenization and global context modeling infrastructure. The backbone is initialized with the same weights pre-trained on large-scale optical--SAR image pairs provided by the TransOSS framework. The novel contributions of SDF-Net are the two physics-guided modules that operate on this base: (1) the SCL module, which introduces a structural consistency loss $\mathcal{L}_{\text{struct}}$ at intermediate layers without adding parameters; and (2) the DFL module, which adds two linear projection heads and an orthogonality loss $\mathcal{L}_{\text{orth}}$ at the terminal stage, followed by parameter-free additive fusion. In the ablation study (Section IV-E), the baseline variant (no SCL, no DFL) uses SDF-Net's own hyperparameter configuration to provide a controlled comparison that isolates the contribution of each module. In the state-of-the-art comparison (Section IV-D), SDF-Net is evaluated alongside the published TransOSS results under their respective tuned settings.

The input image $I$ is initially partitioned into $N_p = (H \times W)/P^2$ non-overlapping patches with a patch stride $P=16$, which are subsequently linearly projected into a sequence of latent tokens. The network processes these tokens through $L$ consecutive Transformer blocks to extract hierarchical features. Let $\mathbf{F}^{(B_s)} \in \mathbb{R}^{B \times C \times H' \times W'}$ denote the feature map derived from the $B_s$-th block. To mitigate the profound modality discrepancy induced by active microwave scattering and passive optical reflectance, SDF-Net departs from purely data-driven black-box alignment. Instead, it introduces two physics-guided components:
\begin{itemize}
    \item Structure-Aware Consistency Learning (SCL): This module is strategically embedded at intermediate feature layers to anchor the network on stable geometric skeletons, effectively decoupling the ship's rigid hull structure from sensor-specific textural fluctuations.
    \item Disentangled Feature Learning (DFL): Operating at the terminal identity stage, this component factorizes the latent representation into a shared identity subspace and a modality-specific auxiliary subspace, employing an additive residual refinement strategy to integrate complementary cues.
\end{itemize}

\subsection{Cross-Modal Feature Tokenization}
As illustrated in Fig.~\ref{fig:framework}(a), to accommodate the disparate radiometric properties of optical and SAR sensors, SDF-Net employs a Cross-modal Dual-head Tokenizer at the input stage. Rather than utilizing a generic patch embedding layer, this specialized tokenizer facilitates a modality-aware projection that acknowledges the unique statistical distributions of each sensor. Specifically, the optical and SAR images are processed via independent linear projection heads, which map the local $P \times P$ patches into a unified $C$-dimensional latent space.

This dual-head configuration is essential for neutralizing low-level sensor discrepancies before the sequences enter the shared Transformer backbone. In the optical domain, the tokenizer primarily captures reflectance-based textural primitives, whereas in the SAR domain, it is tasked with embedding backscatter intensity patterns that are often corrupted by coherent speckle noise. By decoupling the initial tokenization process, the network ensures that the subsequent shared blocks operate on features that have already undergone a coarse radiometric alignment. This preliminary transformation prevents the shared self-attention mechanisms from being dominated by modality-specific intensity biases, thereby allowing the backbone to focus on extracting high-level semantic and geometric invariants.

\subsection{Structure-Aware Consistency Learning (SCL)}

\subsubsection{Intermediate Latent Geometry Excavation}
As depicted in Fig.~\ref{fig:framework}(b), a fundamental premise of SDF-Net is that geometric stability is non-uniformly distributed across network layers. While low-level pixels are excessively corrupted by coherent speckle noise in SAR imagery, and high-level semantic tokens are often too abstract for fine-grained geometric matching, intermediate feature maps act as a critical juncture. These layers retain sufficient spatial topology to characterize the ship's physical layout while being sufficiently abstracted from raw radiometric noise.

Unlike traditional hand-crafted structural descriptors such as HOG or HOPC, that operate directly at the raw pixel level, our approach extracts structural priors from the intermediate latent space. Raw pixel-level gradients are highly susceptible to coherent speckle noise in SAR imagery and high-frequency sea clutter in optical data, inevitably generating severe pseudo-edge artifacts. Conversely, representations at the terminal layers suffer from spatial collapse due to global semantic aggregation. By anchoring our structural probe on the intermediate Transformer feature map $\mathbf{F}^{(B_s)}$, SDF-Net elegantly circumvents both extremes. The preceding self-attention blocks effectively filter out low-level radiometric noise, providing a clean, noise-resilient spatial topology where the true rigid hull contours dominate the gradient field.

To characterize geometric structure independent of modality-specific radiometric responses, we exploit spatial gradient information. For rigid objects such as ships, structural primitives are primarily reflected in spatial intensity variations rather than absolute amplitudes. Gradient operators therefore provide a modality-agnostic descriptor of structural transitions, acting as a high-pass filter that is inherently less sensitive to the multiplicative intensity scaling characteristic of SAR backscatter. For a given intermediate feature map $\mathbf{F} = \mathbf{F}^{(B_s)}$, the first-order partial derivatives are computed via index-shifting to capture bidirectional structural variances:
\begin{equation}
    \begin{split}
        \mathbf{G}_x(h, w) & = \mathbf{F}(h, w+1) - \mathbf{F}(h, w-1), \\
        \mathbf{G}_y(h, w) & = \mathbf{F}(h+1, w) - \mathbf{F}(h-1, w),
    \end{split}
\end{equation}
where $\mathbf{G}_x$ and $\mathbf{G}_y$ characterize the horizontal and vertical gradient fields, respectively.

To derive a holistic structural representation that is robust to local pixel perturbations, we perform a discrete spatial integration (denoted by the $\iint \nabla \mathbf{G}_{x,y}$ operator in Fig. \ref{fig:framework}). By aggregating the absolute gradient magnitudes across the spatial grid, we obtain the structural descriptors:
\begin{equation}
    \begin{split}
        \mathbf{e}_x & = \frac{1}{H'W'} \sum_{h,w} \left| \mathbf{G}_x(h,w) \right|, \\
        \mathbf{e}_y & = \frac{1}{H'W'} \sum_{h,w} \left| \mathbf{G}_y(h,w) \right|.
    \end{split}
\end{equation}
Specifically, given the intermediate feature map $\mathbf{F}^{(B_s)} \in \mathbb{R}^{B \times C \times H' \times W'}$, this spatial integration effectively collapses the spatial dimensions, yielding the gradient energy descriptors $\mathbf{e}_x, \mathbf{e}_y \in \mathbb{R}^{B \times C}$ that summarize the global structural intensity per channel. Crucially, this holistic spatial aggregation mechanism effectively mitigates the adverse impact of isolated high-intensity corner reflectors, or strong scattering points, typical in SAR imagery. Instead of being dominated by discrete, localized peaks that could severely corrupt pixel-to-pixel matching, the aggregated gradient energy captures the macroscopic structural contour of the ship. The resulting integrated descriptor $\mathbf{f}_{\text{struct}} = \mathbf{e}_x + \mathbf{e}_y \in \mathbb{R}^{B \times C}$ serves as a distortion-tolerant, physics-grounded structural anchor. To further neutralize the absolute amplitude disparities between SAR backscatter and optical reflectance, $\mathbf{f}_{\text{struct}}$ undergoes Instance Normalization (IN) applied independently across the channel dimension $C$ for each sample.

\subsubsection{Scale-Invariant Instance Normalization}
The absolute magnitudes of $\mathbf{f}_{\text{struct}}$ remain inherently inconsistent across sensors due to the fundamental divergence in imaging physics. SAR images typically exhibit a high dynamic range and skewed energy distributions due to the corner reflector effects of metallic ship hulls, whereas optical images are governed by diffuse reflectance. To achieve radiometric robustness, we apply Instance Normalization (IN) to the descriptors. This operation functions as a statistical filter that maps the disparate energy distributions into a standardized, unit-variance manifold:
\begin{equation}
    \hat{\mathbf{f}}_{\text{struct}}^{(b)} = \frac{\mathbf{f}_{\text{struct}}^{(b)} - \mu^{(b)}}{\sqrt{(\sigma^{(b)})^2 + \epsilon}},
\end{equation}
where $\mu^{(b)}$ and $\sigma^{(b)}$ are the mean and standard deviation computed over the channel dimension $C$. This normalization effectively strips away modality-specific ``styles''—such as amplitude bias and illumination variance—while preserving the essential geometric ``content'' necessary for cross-modal association.

\subsubsection{Prototype-level Consistency Loss}
To enforce cross-modal coherence, we align the structural descriptors at a stable identity level rather than at an instance level to avoid overfitting to individual sample noise. For each identity $i$ in a mini-batch, we define the modality-specific structural prototypes $\mathbf{c}_i^{o}$ for the optical modality and $\mathbf{c}_i^{s}$ for SAR:
\begin{equation}
    \mathbf{c}_i^{o} = \frac{1}{|\mathcal{B}_i^{o}|} \sum_{b \in \mathcal{B}_i^{o}} \hat{\mathbf{f}}_{\text{struct}}^{(b)}, \quad \mathbf{c}_i^{s} = \frac{1}{|\mathcal{B}_i^{s}|} \sum_{b \in \mathcal{B}_i^{s}} \hat{\mathbf{f}}_{\text{struct}}^{(b)},
\end{equation}
where $\mathcal{B}_i^{o}$ and $\mathcal{B}_i^{s}$ represent the sets of samples for identity $i$ in each modality. The Structure Consistency Loss $\mathcal{L}_{\text{struct}}$ minimizes the Euclidean distance between these prototypes:
\begin{equation}
    \mathcal{L}_{\text{struct}} = \frac{1}{|\mathcal{I}|} \sum_{i \in \mathcal{I}} \left\| \mathbf{c}_i^{o} - \mathbf{c}_i^{s} \right\|_2^2.
\end{equation}
This loss imposes a strong geometric prior on the network, forcing it to prioritize modality-invariant geometric signatures over transient and unreliable textural features.

In practice, the intermediate feature map is extracted from the $B_s$-th Transformer block.
We empirically select $B_s=6$ for a 12-layer backbone,
as this layer balances spatial detail preservation and semantic abstraction.
Earlier layers tend to be dominated by low-level noise (especially SAR speckle),
while deeper layers become overly identity-focused and lose fine-grained structural information.
Ablation experiments in Sec.~IV further validate the robustness of this choice.

\subsection{Disentangled Feature Learning and Residual Fusion}

As shown in Fig.~\ref{fig:framework}(c), at the terminal stage, the abstract representation $\mathbf{F}^{(L)} \in \mathbb{R}^{B \times d}$ is fed into two independent linear projection heads, implemented as fully-connected layers without non-linear activations. These parallel branches factorize the terminal representation into the shared identity subspace $\mathbf{f}_{\text{sh}} \in \mathbb{R}^{B \times d}$ and the modality-specific subspace $\mathbf{f}_{\text{sp}} \in \mathbb{R}^{B \times d}$. By maintaining the identical dimensionality $d$ for both subspaces, we facilitate the subsequent element-wise additive residual fusion without requiring additional channel-matching convolutions. Inspired by the success of orthogonal subspace projection in optimizing deep representation learning \cite{sun2017svdnet}, we impose an Orthogonality Constraint to ensure the mathematical independence of these feature subspaces:
\begin{equation}
    \mathcal{L}_{\text{orth}} = \mathbb{E} \left[ \left| \left\langle \bar{\mathbf{f}}_{\text{sh}}, \bar{\mathbf{f}}_{\text{sp}} \right\rangle \right| \right],
\end{equation}
where $\bar{\mathbf{f}}$ denotes $\ell_2$-normalized features. By mathematically minimizing the mutual information between these subspaces, the network is forced to rigorously isolate sensor-independent identity cues in $\mathbf{f}_{\text{sh}}$, effectively preventing modality-unique textural nuances from contaminating the invariant geometric anchor.

Rather than discarding $\mathbf{f}_{\text{sp}}$ as noise, we integrate these features via an additive complementarity strategy:
\begin{equation}
    \mathbf{f}_{\text{fuse}} = \mathbf{f}_{\text{sh}} + \mathbf{f}_{\text{sp}}.
\end{equation}
Physically, this element-wise addition operates as a robust residual refinement mechanism. Since the shared feature $\mathbf{f}_{\text{sh}}$ is strictly regularized by the geometric consistency loss $\mathcal{L}_{\text{struct}}$ to act as the primary cross-modal anchor, superimposing $\mathbf{f}_{\text{sp}}$ does not destruct its established modality invariance. Instead, the specific feature $\mathbf{f}_{\text{sp}}$ adaptively supplements fine-grained, sensor-dependent identity nuances such as unique superstructure scattering distributions or distinct paint reflectances that are vital for differentiating highly similar ships, thereby maximizing the ultimate re-identification precision without introducing dimensional redundancy.

\subsection{Joint Optimization Objective}
The unified loss function for training SDF-Net is defined as:
\begin{equation}
    \mathcal{L} = \mathcal{L}_{\text{id}} + \lambda_{\text{orth}} \mathcal{L}_{\text{orth}} + \lambda_{\text{struct}} \mathcal{L}_{\text{struct}},
\end{equation}
where $\mathcal{L}_{\text{id}}$ encompasses the label-smoothed cross-entropy loss and the weighted triplet loss for identity supervision. The hyper-parameters $\lambda_{\text{orth}}$ and $\lambda_{\text{struct}}$ serve to balance the influence of feature disentanglement and structural consistency. Through this joint optimization, SDF-Net achieves a harmonious convergence toward a physics-informed identity space that is both robust to modality shifts and highly discriminative for maritime surveillance.

To provide a clear and holistic perspective of the feature abstraction mechanism, the complete forward processing pipeline of SDF-Net is summarized in Algorithm \ref{alg:sdfnet_forward}. This algorithmic formulation delineates how an input image progressively evolves through the dual-head tokenizer, intermediate structural extraction, and terminal feature disentanglement to form the final robust representation.

\begin{algorithm}[htbp]
    \caption{Forward Processing Pipeline of SDF-Net}
    \label{alg:sdfnet_forward}
    \begin{algorithmic}
        \REQUIRE Input image $I$, modality $m \in \{\text{optical}, \text{SAR}\}$, layer index $B_s$, total layers $L$.
        \ENSURE Final discriminative identity representation $\mathbf{f}_{\text{fuse}}$.

        \STATE 1. Cross-modal Dual-head Tokenization
        \IF{$m = \text{optical}$}
        \STATE $\mathbf{T} \leftarrow \text{Tokenizer}_{opt}(I)$
        \ELSE
        \STATE $\mathbf{T} \leftarrow \text{Tokenizer}_{sar}(I)$
        \ENDIF

        \STATE 2. Intermediate Feature Extraction
        \STATE $\mathbf{F}^{(B_s)} \leftarrow \text{TransformerBlocks}_{1 \to B_s}(\mathbf{T})$

        \STATE 3. Structure-Aware Consistency Learning (SCL)
        \STATE $\mathbf{G}_x, \mathbf{G}_y \leftarrow \nabla_{x,y} \mathbf{F}^{(B_s)}$
        \STATE $\mathbf{f}_{\text{struct}} \leftarrow \iint (|\mathbf{G}_x| + |\mathbf{G}_y|)$
        \STATE $\hat{\mathbf{f}}_{\text{struct}} \leftarrow \text{IN}(\mathbf{f}_{\text{struct}})$
        \STATE \textit{/* Training constraint: Optimize $\mathcal{L}_{\text{struct}}$ across modalities */}

        \STATE 4. Terminal Feature Extraction
        \STATE $\mathbf{F}^{(L)} \leftarrow \text{TransformerBlocks}_{B_s+1 \to L}(\mathbf{F}^{(B_s)})$

        \STATE 5. Disentangled Feature Learning (DFL)
        \STATE $\mathbf{f}_{\text{sh}}, \mathbf{f}_{\text{sp}} \leftarrow \text{Decouple}(\mathbf{F}^{(L)})$
        \STATE \textit{/* Training constraint: Optimize $\mathcal{L}_{\text{orth}} = \mathbb{E}[|\langle \bar{\mathbf{f}}_{\text{sh}}, \bar{\mathbf{f}}_{\text{sp}} \rangle|]$ */}

        \STATE 6. Additive Residual Fusion
        \STATE $\mathbf{f}_{\text{fuse}} \leftarrow \mathbf{f}_{\text{sh}} + \mathbf{f}_{\text{sp}}$

        \RETURN $\mathbf{f}_{\text{fuse}}$
    \end{algorithmic}
\end{algorithm}

\section{Experiments}

\subsection{Dataset}

Evaluations are conducted on the HOSS-ReID benchmark, which serves as the primary publicly available resource specifically curated for the cross-modal association of maritime targets. Following the standard person/vehicle ReID protocol such as Market-1501, HOSS-ReID provides tight instance-level bounding-box crops of individual ships extracted from satellite scenes, rather than full-scene imagery. Each crop contains only the ship itself against a near-uniform maritime background; complex environmental factors such as sea clutter, wave patterns, and wake features are largely excluded at the crop level. This dataset design reflects the practical deployment paradigm of a two-stage pipeline: a ship detector first localizes and extracts individual vessel instances from full-scene optical/SAR imagery, and the ReID model then matches identities across the resulting crops. The dataset provides the requisite complexity to validate the robustness of SDF-Net under extreme radiometric disparities between active and passive sensing manifolds. While contemporary maritime datasets such as CMShipReID focus on the alignment of passive sensors including visible, near-infrared, and thermal infrared modalities, HOSS-ReID facilitates research in the more challenging optical--SAR domain where targets are subjected to coherent speckle noise and radar-specific geometric distortions.

The optical imagery was acquired by the Jilin-1 optical constellation at a ground sampling distance of 0.75~m, producing 8-bit RGB images. The SAR imagery was acquired by the TY-MINISAR SAR constellation at 1.0~m, producing single-channel 32-bit floating-point images. Both constellations performed coordinated multi-angle imaging of high-density maritime regions including the Panama Canal and Suez Canal, within short time windows, capturing the same ship from multiple viewing geometries across minutes to days. All images were preprocessed with geometric and radiometric correction; no orthorectification based on a digital elevation model was applied, preserving the native sensor geometry.

The 13 raw image sequences, totaling 43 large-format frames, were manually annotated: ship instances were delineated with bounding boxes and extracted as individual crops. Cross-modal identity association, matching the same physical ship across its optical and SAR appearances, was performed by human annotators using spatial proximity, temporal continuity, and visual inspection of ship characteristics---size, shape, and superstructure layout---to establish ground-truth identity labels. This annotation protocol yields 449 distinct ship trajectories, from which the training and testing splits are derived following the open-set ReID protocol described above.

\begin{table}[htbp]
    \centering
    \caption{Detailed statistical distribution of the HOSS-ReID testing set across different evaluation protocols.}
    \label{tab:hoss_stats}
    \begin{tabular}{ccccc}
        \toprule
        Protocol                        & Modality          & Query images & Gallery images & Total        \\ \midrule
        \multirow{3}{*}{All-to-All}     & Optical           & 88           & 403            & 491          \\
                                        & SAR               & 88           & 190            & 278          \\
                                        & \textit{Subtotal} & \textbf{176} & \textbf{593}   & \textbf{769} \\ \midrule
        \multirow{3}{*}{Optical-to-SAR} & Optical           & 65           & 0              & 65           \\
                                        & SAR               & 0            & 190            & 190          \\
                                        & \textit{Subtotal} & \textbf{65}  & \textbf{190}   & \textbf{255} \\ \midrule
        \multirow{3}{*}{SAR-to-Optical} & SAR               & 67           & 0              & 67           \\
                                        & Optical           & 0            & 403            & 403          \\
                                        & \textit{Subtotal} & \textbf{67}  & \textbf{403}   & \textbf{470} \\ \bottomrule
    \end{tabular}
\end{table}

The training partition consists of 1,063 images in total, encompassing 574 optical and 489 SAR instances, covering 361 distinct ship identities. Following the standard open-set ReID protocol established by Market-1501, the training and testing identities are completely disjoint---no ship identity appears in both partitions. This ensures that evaluation measures cross-modal identity \emph{generalization} to unseen ships rather than memorization of training identities. The testing set comprises 88 query identities (represented by 176 query images, split evenly between optical and SAR) and 251 gallery identities comprising 593 images, of which 163 are distractor identities that appear exclusively in the gallery. These distractors, which have no corresponding query, simulate the realistic open-world scenario where a retrieval system must discriminate target ships from a large pool of unknown vessels. Such a distribution facilitates the network in learning latent embeddings that remain stable despite the presence of non-linear radiometric fluctuations and varying maritime environmental conditions. Transitioning to the evaluation phase, the testing set is organized into three distinct retrieval protocols to quantify bidirectional and uni-directional search performance. Specifically, Optical-to-SAR uses optical images as queries to retrieve matching targets from a SAR gallery, simulating the real-world scenario of searching for a visually identified ship within historical radar records. Conversely, SAR-to-Optical performs the reverse, using SAR queries to search an optical gallery, which is critical for identifying targets detected at night or under cloud cover. Finally, the All-to-All protocol combines all queries and gallery images regardless of modality, providing a comprehensive and holistic measure of modality-invariant alignment. The precise statistical breakdown of the query and gallery compositions for each protocol is detailed in Table \ref{tab:hoss_stats}.

\subsection{Evaluation Metrics}

To rigorously quantify the cross-modal retrieval performance of SDF-Net, we employ two categories of standardized metrics widely recognized in the re-identification community \cite{pami21reidsurvey}. The Cumulative Match Characteristic (CMC) is utilized to assess the identity matching capability, where the Rank-$k$ accuracy represents the probability that at least one correctly matched candidate appears within the top-$k$ retrieved results. We specifically report Rank-1, Rank-5, and Rank-10 scores to evaluate the model's precision across varying retrieval breadths. While Rank-1 reflects the primary identification accuracy, the higher-rank metrics provide insight into the robustness of the learned manifold in preserving identity proximity despite severe radiometric distortions.

Parallel to the ranking accuracy, the mean Average Precision (mAP) serves as a holistic descriptor of the retrieval efficacy by accounting for both precision and recall across the entire gallery. For a given query, the Average Precision (AP) is calculated by integrating the area under the precision--recall curve, formulated as:
\begin{equation}
    AP = \frac{\sum_{n=1}^{N} P(n) \times \text{rel}(n)}{G},
\end{equation}
where $N$ denotes the total number of images in the gallery, $G$ is the count of ground-truth matches, and $P(n)$ represents the precision at the $n$-th rank. The indicator function $\text{rel}(n)$ is unity if the $n$-th result is a correct match and zero otherwise. The final mAP is derived by averaging the AP values across the entire query set of size $Q$:
\begin{equation}
    mAP = \frac{\sum_{q=1}^{Q} AP_q}{Q}.
\end{equation}
In the context of optical--SAR ship re-identification, mAP is particularly critical as it penalizes the failure to retrieve all instances of a target ship, thereby ensuring that the structural anchors learned by SDF-Net effectively bridge the modality gap for all samples of the same identity. These metrics collectively provide a multidimensional assessment of the model's ability to maintain high discriminative power within a volatile maritime sensing environment.

\subsection{Implementation Details}
All experiments were conducted on a single NVIDIA RTX 3090 GPU equipped with 24GB of VRAM. The software environment is built upon PyTorch 2.2.2 and CUDA 11.8 for hardware acceleration. We adopt the base variant of the Vision Transformer (ViT-B/16) as our backbone network, initialized with weights pre-trained on optical--SAR paired data provided by the TransOSS framework. The input images are uniformly resized to $256 \times 128$ pixels. To augment the training data and prevent overfitting, we apply random horizontal flipping with a probability of 0.5, random cropping with zero padding, and random erasing with a probability of 0.2. The backbone is initialized with ViT-B/16 weights pre-trained on large-scale optical--SAR image pairs from the TransOSS framework. In the ablation study (Section IV-E), all four variants (baseline, SCL-only, DFL-only, and full SDF-Net) share the same hyperparameter configuration to ensure a controlled component-wise comparison; the only difference is whether SCL and DFL are enabled.

To guarantee the computability of the cross-modal prototype consistency loss $\mathcal{L}_{\text{struct}}$ and avoid single-modality batches, we implement a strict cross-modal $P \times K$ sampling strategy. Specifically, each mini-batch of size 32 is constrained to contain exactly $P=8$ distinct ship identities. For every identity, we randomly sample $K=4$ instances, strictly ensuring a balanced composition of 2 optical images and 2 SAR images. The network is optimized using the Stochastic Gradient Descent (SGD) optimizer with a weight decay of $1 \times 10^{-4}$. The initial base learning rate is set to $5 \times 10^{-4}$, which incorporates a linear warmup strategy in the early stage before smoothly decaying over a total of 100 training epochs. The hyper-parameters for the joint loss optimization are empirically set to $\lambda_{\text{orth}} = 10.0$ and $\lambda_{\text{struct}} = 1.0$, with the intermediate structural features extracted from the $B_s=6$ Transformer block.

\subsection{Comparison with State-of-the-Art Methods}

The performance of SDF-Net is evaluated against a diverse set of representative algorithms on the HOSS-ReID benchmark, with the comparative results summarized in Table \ref{tab:sota}. The baseline and state-of-the-art methods are categorized into general vision backbones, single-modality re-identification models, and cross-modal retrieval frameworks to provide a comprehensive assessment of the current landscape in maritime surveillance.

General vision backbones and single-modality re-identification models exhibit a pronounced susceptibility to the severe non-linear radiometric distortions inherent in optical--SAR imagery. Although transformer-based architectures such as DeiT-base and TransReID demonstrate superior feature extraction capabilities compared to earlier convolutional counterparts, their performance remains suboptimal due to the absence of explicit cross-modal alignment mechanisms. For instance, TransReID achieves a modest 20.9\% mAP under the SAR-to-Optical protocol, underscoring the inadequacy of standard appearance-based matching when confronted with the coherent speckle noise and geometric artifacts of radar imaging. While these models effectively capture high-level semantics, they fail to bridge the disparate sensing manifolds without specialized constraints.

Existing cross-modal re-identification methods, primarily optimized for visible--infrared person re-identification, similarly struggle to generalize to the maritime domain. Approaches such as DEEN and VersReID are designed to align passive thermal radiation with visible reflectance, a task that does not account for the drastic imaging discrepancies between active microwave backscattering and optical imagery. This domain mismatch is evident in the performance of AMML and CM-NAS, which yield significantly lower accuracies than the remote-sensing-specific baseline. The inability of these models to capture stable geometric invariants leads to substantial alignment failures in the presence of fluctuating sea clutter and varying draft depths.

\begin{table*}[t]
    \centering
    \caption{Quantitative evaluation of the proposed SDF-Net compared with state-of-the-art methods on the HOSS-ReID benchmark. All values are reported in percentage (\%).}
    \setlength{\tabcolsep}{5pt}
    \renewcommand{\arrayrulewidth}{1pt}
    \begin{tabular}{ccccccccccccccc}
        \toprule
        \multirow{2.5}{*}{Task Type}          & \multirow{2.5}{*}{Method}            & \multirow{2.5}{*}{Venue} & \multicolumn{4}{c}{All-to-All} & \multicolumn{4}{c}{Optical-to-SAR} & \multicolumn{4}{c}{SAR-to-Optical}                                                                                                                                                 \\ \cmidrule(lr){4-7}\cmidrule(lr){8-11}\cmidrule(lr){12-15}
                                              &                                      &                          & mAP                            & R1                                 & R5                                 & R10           & mAP           & R1            & R5            & R10           & mAP           & R1            & R5            & R10           \\
        \midrule
        \multirow{2}{*}{General model}        & ViT-base~\cite{dosovitskiy2020image} & Arxiv2020                & 43.0                           & 56.2                               & 64.8                               & 69.9          & 21.5          & 12.3          & 33.8          & 55.4          & 17.9          & 10.4          & 25.4          & 32.8          \\
                                              & DeiT-base~\cite{touvron2021training} & ICML2021                 & 47.2                           & 58.1                               & 69.6                               & 74.1          & 25.9          & 16.1          & 36.7          & 59.1          & 26.1          & 10.3          & 35.7          & 52.0          \\
        \midrule
        \multirow{4}{*}{Single modality ReID} & AGW~\cite{pami21reidsurvey}          & TPAMI2021                & 43.6                           & 57.4                               & 64.2                               & 68.8          & 17.2          & 7.7           & 29.2          & 38.5          & 21.1          & 14.9          & 34.3          & 46.3          \\
                                              & TransReID~\cite{he2021transreid}     & ICCV2021                 & 48.1                           & 60.8                               & 69.3                               & 73.9          & 27.3          & 18.5          & 40.0          & 58.5          & 20.9          & 11.9          & 34.3          & 43.3          \\
                                              & SOLIDER~\cite{chen2023beyond}        & CVPR2023                 & 38.2                           & 50.6                               & 63.1                               & 69.9          & 23.1          & 12.3          & 38.5          & 52.3          & 14.6          & 10.4          & 16.4          & 31.3          \\
                                              & D2InterNet~\cite{liu2025advancing}   & SIGIR2025                & 50.2                           & 59.1                               & 71.6                               & 79.0          & 33.0          & 21.5          & 41.5          & 69.8          & 28.8          & 25.4          & 38.8          & 50.7          \\
        \midrule
        \multirow{10}{*}{Cross-modal ReID}    & Hc-Tri~\cite{liu2020parameter}       & TMM2020                  & 34.0                           & 47.2                               & 54.6                               & 59.7          & 11.1          & 6.2           & 15.4          & 24.6          & 10.9          & 7.5           & 20.9          & 29.9          \\
                                              & CM-NAS~\cite{fu2021cm}               & CVPR2021                 & 30.7                           & 46.0                               & 54.6                               & 57.4          & 8.2           & 1.5           & 10.8          & 21.5          & 7.6           & 4.5           & 11.9          & 19.4          \\
                                              & LbA~\cite{park2021learning}          & CVPR2021                 & 33.0                           & 48.3                               & 59.7                               & 62.5          & 11.9          & 4.6           & 23.1          & 41.5          & 8.5           & 6.0           & 14.9          & 22.4          \\
                                              & DEEN~\cite{zhang2023diverse}         & CVPR2023                 & 43.8                           & 58.5                               & 64.2                               & 66.5          & 31.3          & 21.5          & 44.6          & 60.0          & 27.4          & 22.4          & 40.3          & 53.7          \\
                                              & MCJA~\cite{liang2024bridging}        & TCSVT2024                & 47.1                           & 59.1                               & 67.9                               & 73.0          & 18.6          & 10.8          & 27.7          & 38.5          & 19.7          & 14.9          & 28.3          & 43.3          \\
                                              & VersReID~\cite{zheng2024versatile}   & TPAMI2024                & 49.3                           & 59.7                               & 70.5                               & 78.4          & 25.7          & 13.8          & 40.0          & 61.5          & 27.7          & 17.9          & 44.8          & 61.2          \\
                                              & AMML~\cite{zhang2025adaptive}        & IJCV2025                 & 31.2                           & 43.8                               & 52.8                               & 56.8          & 9.1           & 4.6           & 10.8          & 21.5          & 9.2           & 4.5           & 13.4          & 20.9          \\
                                              & HSFLNet~\cite{zhu2025hypergraph}     & EAAI2025                 & 22.8                           & 29.0                               & 44.9                               & 52.3          & 19.0          & 13.9          & 24.6          & 32.3          & 19.2          & 19.4          & 26.9          & 46.3          \\
                                              & TransOSS~\cite{Wang_2025_ICCV}       & ICCV2025                 & 57.4                           & 65.9                               & 79.5                               & 85.8          & 48.9          & 33.8          & \textbf{67.7} & 80.0          & 38.7          & 29.9          & 59.7          & 71.6          \\
                                              & \textbf{SDF-Net (ours)}              & -                        & \textbf{60.9}                  & \textbf{69.9}                      & \textbf{81.8}                      & \textbf{88.1} & \textbf{50.0} & \textbf{35.4} & \textbf{67.7} & \textbf{86.2} & \textbf{46.6} & \textbf{38.8} & \textbf{70.1} & \textbf{76.1} \\
        \bottomrule
    \end{tabular}
    \label{tab:sota}
\end{table*}

SDF-Net consistently achieves highly competitive or superior performance over all state-of-the-art methods across every evaluation metric and retrieval protocol. Under the comprehensive All protocol, the proposed method achieves 60.9\% mAP and 69.9\% Rank-1 accuracy, representing an improvement of 3.5\% and 4.0\% respectively over the current leading baseline, TransOSS. The performance margin is particularly significant in the SAR-to-Optical task, where SDF-Net elevates the mAP from 38.7\% to 46.6\%. This 7.9\% absolute increase validates the efficacy of anchoring identity representations on modality-invariant geometric skeletons through the Structure-Aware Consistency learning module. By leveraging the rigid hull structure as a definitive physical anchor, SDF-Net maintains a robust shared manifold that is far less sensitive to radiometric fluctuations than the implicit global attention mechanisms utilized in TransOSS. The additive residual fusion strategy further enhances the discriminative precision of the model by integrating modality-specific nuances as refinements rather than noise, establishing a new state-of-the-art for cross-modal maritime target association.

\subsection{Ablation Studies}

The architectural integrity and performance gains of SDF-Net are validated through a systematic dissection across three critical axes. First, we evaluate the fundamental contributions of the Structure-Aware Consistency Learning (SCL) and Disentangled Feature Learning (DFL) modules to establish the necessity of each component. Second, the efficacy of disparate feature integration strategies—comprising summation, shared-only, specific-only, and concatenation—is assessed to determine the optimal fusion paradigm. Finally, we investigate the sensitivity of the geometric extraction layer to identify the block index that most effectively captures modality-invariant structural primitives. All experiments are conducted on the HOSS-ReID dataset following the standard evaluation protocols.

\subsubsection{Effectiveness of Proposed Modules}

The individual and synergistic impacts of SCL and DFL are quantified by progressively integrating them into the vanilla backbone. As evidenced by the results in Table \ref{tab:ablation_modules}, the isolated inclusion of SCL yields a measurable enhancement in retrieval stability, particularly elevating the SAR-to-Optical mAP from 44.5\% to 46.6\%. This improvement suggests that enforcing geometric consistency at intermediate layers effectively anchors the representation space against the radiometric fluctuations inherent in SAR imagery. Notably, while the SCL constraint primarily optimizes the structural alignment, the DFL module serves as a decisive driver for discriminative precision. The introduction of DFL significantly boosts the Rank-1 accuracy from 67.6\% to 69.9\% in the All-to-All setting, confirming that decoupling modality-specific noise from shared identity cues facilitates a more robust feature manifold.

\begin{table}[htbp]
    \centering
    \caption{Ablation study highlighting the effectiveness of the SCL and DFL modules on the HOSS-ReID benchmark. All values are reported in percentage (\%).}
    \label{tab:ablation_modules}
    \begin{tabular}{cccccccc}
        \toprule
        \multirow{2}{*}{SCL} & \multirow{2}{*}{DFL} & \multicolumn{2}{c}{All-to-All} & \multicolumn{2}{c}{Optical-to-SAR} & \multicolumn{2}{c}{SAR-to-Optical}                                                 \\ \cmidrule(lr){3-4}\cmidrule(lr){5-6}\cmidrule(lr){7-8}
                             &                      & mAP                            & R1                                 & mAP                                & R1            & mAP           & R1            \\ \midrule
        \ding{55}            & \ding{55}            & 58.6                           & 67.6                               & 46.5                               & 32.3          & 44.5          & \textbf{38.8} \\
        \ding{51}            & \ding{55}            & 59.2                           & 66.5                               & 47.6                               & 32.3          & \textbf{46.6} & 37.3          \\
        \ding{55}            & \ding{51}            & 59.8                           & \textbf{69.9}                      & 49.3                               & \textbf{35.4} & 41.4          & 31.3          \\
        \ding{51}            & \ding{51}            & \textbf{60.9}                  & \textbf{69.9}                      & \textbf{50.0}                      & \textbf{35.4} & \textbf{46.6} & \textbf{38.8} \\ \bottomrule
    \end{tabular}
\end{table}

Table~\ref{tab:ablation_modules} reveals three notable patterns. SCL applied alone improves SAR-to-Optical mAP by 2.1\%, from 44.5\% to 46.6\%, but decreases Rank-1 by 1.5\%, from 38.8\% to 37.3\%. This trade-off has a physical interpretation: the structure consistency loss $\mathcal{L}_{\text{struct}}$ steers the network toward geometric features---hull contour and aspect ratio---and away from modality-specific texture. Geometry is stable across sensors, which produces more correct matches spread throughout the ranked list and thus a higher mAP, but it is inherently less discriminative for visually similar ships, so the single top-ranked result can occasionally be a geometrically similar but incorrect identity, yielding a lower Rank-1. This is the classic robustness--discriminability trade-off: what SCL gains in cross-modal stability, it partially sacrifices in fine-grained precision.

DFL applied alone degrades SAR-to-Optical performance substantially (mAP from 44.5\% to 41.4\%, Rank-1 from 38.8\% to 31.3\%). The orthogonality constraint $\mathcal{L}_{\text{orth}}$ alone, without the geometric anchor provided by SCL, cannot prevent the modality-specific subspaces $\mathbf{f}_{\text{sp}}$ from drifting into mutually incompatible regions. In the SAR-to-Optical protocol, SAR queries and optical galleries occupy disjoint modality spaces; without geometric regularization, the disentangled features of the two modalities become unalignable. This confirms that disentanglement without structural guidance is harmful in the active--passive cross-modal setting.

The combination of SCL and DFL achieves the best results across all metrics: SAR-to-Optical mAP 46.6\% (tying SCL-only) and Rank-1 38.8\%, restoring the baseline level, with All-to-All mAP and Rank-1 reaching 60.9\% and 69.9\%, respectively. SCL provides the geometric anchor that stabilizes the shared representation space, enabling DFL to refine it with modality-specific residual information without the subspace drift observed in DFL-only. SCL and DFL are not independent additive modules but complementary components whose combination is essential for robust discriminability: the geometric regularization of SCL creates the conditions under which DFL's disentanglement becomes beneficial rather than harmful.

\subsubsection{Sensitivity Analysis of Structural Extraction Layer}

To determine the optimal abstraction level for anchoring geometric consistency, we investigate the impact of the insertion depth for the Structure Consistency Constraint by varying the feature extraction layer index $B_s$ within the Vision Transformer backbone. As presented in Table \ref{tab:ablation_layer}, extracting structural priors from shallow layers such as $B_s=2$ yields suboptimal retrieval accuracy. This underperformance is attributed to the prevalence of low-level radiometric noise and coherent speckle artifacts in the early processing stages, which corrupt the gradient energy statistics before sufficient semantic filtering occurs.

Performance metrics improve significantly as the extraction point shifts toward intermediate layers, confirming that mid-level representations effectively retain spatial topology while abstracting away sensor-specific interference. Specifically, the configuration with $B_s=6$ attains the peak holistic performance, achieving an mAP of 60.9\% and Rank-1 accuracy of 69.9\% under the All protocol. Although $B_s=4$ exhibits competitive results and slightly higher Optical-to-SAR precision, the $B_s=6$ setting demonstrates superior robustness in the more challenging SAR-to-Optical scenario, delivering an mAP of 46.6\% compared to 45.3\% at $B_s=4$. This indicates that the sixth layer offers a more balanced trade-off between suppressing optical texture variations and preserving radar geometric signatures.

Conversely, enforcing structural consistency at deeper layers leads to a discernible performance degradation. The results at $B_s=8$ and $B_s=10$ reveal that high-level semantic features become overly abstract and spatially collapsed, thereby losing the fine-grained geometric layout information essential for pixel-wise gradient alignment. Consequently, intermediate layers serve as the most reliable structural probe, motivating the selection of $B_s=6$ as the default configuration for SDF-Net.

\begin{table}[htbp]
    \centering
    \caption{Ablation study on the structural extraction layer index $B_s$. The results indicate that extracting structural priors at intermediate layers such as $B_s=6$, achieves the optimal trade-off between spatial fidelity and semantic abstraction. All values are reported in percentage (\%).}
    \label{tab:ablation_layer}
    % 使用 booktabs 宏包画线
    \renewcommand{\arrayrulewidth}{1pt}
    \setlength{\tabcolsep}{3.5mm}
    \begin{tabular}{ccccccc}
        \toprule
        \multirow{2}{*}{$B_s$} & \multicolumn{2}{c}{All-to-All} & \multicolumn{2}{c}{Optical-to-SAR} & \multicolumn{2}{c}{SAR-to-Optical}                                                 \\
        \cmidrule(lr){2-3}\cmidrule(lr){4-5}\cmidrule(lr){6-7}
                               & mAP                            & R1                                 & mAP                                & R1            & mAP           & R1            \\
        \midrule
        2                      & 59.7                           & 68.2                               & 49.0                               & 35.4          & 46.0          & \textbf{40.3} \\
        4                      & 60.4                           & 68.8                               & \textbf{50.5}                      & \textbf{38.5} & 45.3          & 38.8          \\
        \textbf{6}             & \textbf{60.9}                  & \textbf{69.9}                      & 50.0                               & 35.4          & \textbf{46.6} & 38.8          \\
        8                      & 58.4                           & 65.3                               & 48.7                               & 35.4          & 45.5          & \textbf{40.3} \\
        10                     & 58.7                           & 66.5                               & 48.9                               & 33.8          & 44.7          & 37.3          \\
        12                     & 60.3                           & \textbf{69.9}                      & 47.4                               & 33.8          & 45.3          & 34.3          \\
        \bottomrule
    \end{tabular}
\end{table}

\subsubsection{Evaluation of Feature Fusion Strategies}

To validate the rationale behind the additive complementarity design, we examine the discriminative contribution of the disentangled subspaces and compare different integration mechanisms. As detailed in Table \ref{tab:ablation_fusion}, utilizing the modality-specific feature $\mathbf{f}_{\text{sp}}$ in isolation yields the lowest performance, with an mAP of 58.7\% under the All protocol. This confirms that sensor-dependent characteristics, such as SAR speckle patterns or optical color textures, are insufficient for reliable cross-modal matching when detached from the underlying identity structure.

In contrast, the shared identity feature $\mathbf{f}_{\text{sh}}$ alone achieves a reputable mAP of 59.2\%, substantiating the effectiveness of the orthogonality constraint in isolating modality-invariant cues. However, a performance plateau is observed, indicating that discarding all modality-specific information inevitably results in a loss of fine-grained discriminative details.

Integrating these two subspaces leads to further improvements, yet the fusion method proves critical. The concatenation strategy $\text{Cat}(\mathbf{f}_{\text{sh}}, \mathbf{f}_{\text{sp}})$ results in an mAP of 59.5\% under the All protocol. Although it slightly edges out the additive strategy in the Optical-to-SAR mAP, at 50.7\%, it introduces a twofold expansion of the feature dimension, imposing unnecessary computational overhead and redundancy on the subsequent retrieval heads. Conversely, the proposed additive fusion $\mathbf{f}_{\text{sh}} + \mathbf{f}_{\text{sp}}$ delivers the most comprehensive and robust performance, reaching the peak 60.9\% mAP and 69.9\% Rank-1 accuracy under the primary All-to-All protocol, while also dominating the more challenging SAR-to-Optical task. This suggests that treating modality-specific features as a parameter-efficient residual refinement to the shared identity embedding effectively maximizes discriminability without expanding the original feature dimensionality.

\begin{table}[htbp]
    \centering
    \caption{Ablation study on feature fusion strategies. The additive interaction acts as a parameter-efficient residual refinement, outperforming concatenation and single-branch baselines. All values are reported in percentage (\%).}
    \label{tab:ablation_fusion}
    \footnotesize
    \setlength{\tabcolsep}{3.5pt}
    \renewcommand{\arrayrulewidth}{1pt}
    \begin{tabular}{ccccccc}
        \toprule
        \multirow{2}{*}{Fusion Strategy}         & \multicolumn{2}{c}{All-to-All} & \multicolumn{2}{c}{Optical-to-SAR} & \multicolumn{2}{c}{SAR-to-Optical}                                                 \\
        \cmidrule(lr){2-3}\cmidrule(lr){4-5}\cmidrule(lr){6-7}
                                                 & mAP                            & R1                                 & mAP                                & R1            & mAP           & R1            \\
        \midrule
        Specific-only ($\mathbf{f}_{\text{sp}}$) & 58.7                           & 67.6                               & 44.3                               & 29.2          & 43.9          & 35.8          \\
        Shared-only ($\mathbf{f}_{\text{sh}}$)   & 59.2                           & 68.2                               & 49.8                               & 33.8          & 43.1          & 37.3          \\
        Concatenation                            & 59.5                           & 68.8                               & \textbf{50.7}                      & \textbf{35.4} & 45.1          & 35.8          \\
        \textbf{Additive (Ours)}                 & \textbf{60.9}                  & \textbf{69.9}                      & 50.0                               & \textbf{35.4} & \textbf{46.6} & \textbf{38.8} \\
        \bottomrule
    \end{tabular}
\end{table}

\subsubsection{Computational Complexity Analysis}

To validate the efficiency of the proposed framework, we quantify the computational complexity and parameter count of SDF-Net against the leading baseline, TransOSS. As summarized in Table \ref{tab:complexity}, thanks to the parameter-free nature of the spatial gradient integration and instance normalization within the SCL module, alongside the element-wise additive fusion in the DFL module, SDF-Net introduces absolutely zero additional parameters. Specifically, both SDF-Net and the baseline strictly maintain 86.24 M parameters. In terms of computational overhead, SDF-Net requires 22.42 G FLOPs during inference, representing a negligible increase of merely 0.17 G FLOPs compared to the 22.25 G FLOPs of the baseline. This marginal computational addition, less than a 0.8\% increase, yields a substantial 3.5\% absolute improvement in the comprehensive mAP, from 57.4\% to 60.9\%, and a 4.0\% increase in Rank-1 accuracy, from 65.9\% to 69.9\%. These statistics conclusively demonstrate that SDF-Net achieves superior physics-guided cross-modal alignment in a highly parameter- and compute-efficient manner, avoiding the prohibitive overhead typical of complex generative or attention-heavy alignment strategies.

\begin{table}[htbp]
    \centering
    \caption{Computational complexity and performance comparison between the baseline and the proposed SDF-Net.}
    \label{tab:complexity}
    \renewcommand{\arrayrulewidth}{1pt}
    \begin{tabular}{ccccc}
        \toprule
        Method & Params (M) & FLOPs (G) & mAP (\%) & Rank-1 (\%) \\
        \midrule
        TransOSS & 86.24 & 22.25 & 57.4 & 65.9 \\
        \textbf{SDF-Net} & 86.24 & 22.42 & 60.9 & 69.9 \\
        \bottomrule
    \end{tabular}
\end{table}

\subsubsection{Hyper-parameter Sensitivity Analysis}

\begin{figure}[htbp]
    \centering
    \includegraphics[width=\linewidth]{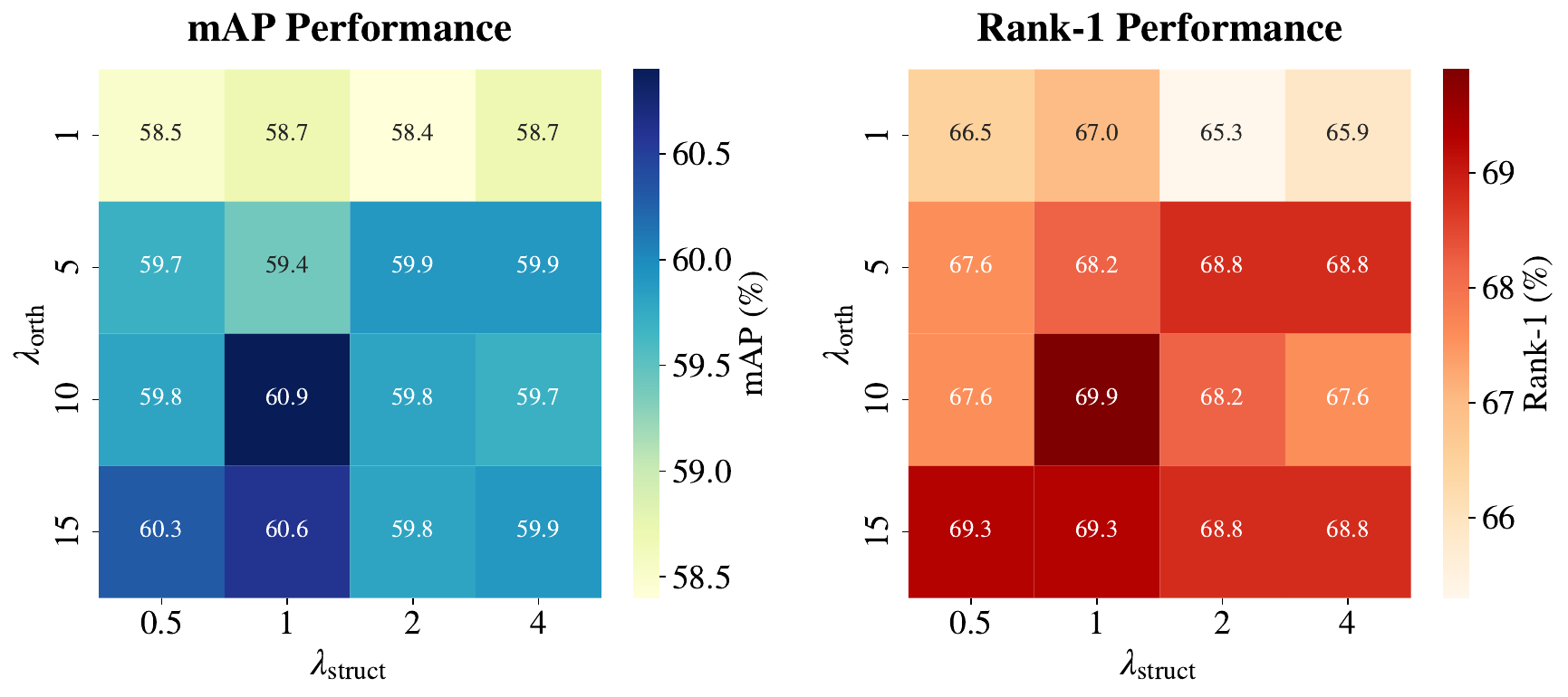}
    \caption{Hyper-parameter sensitivity analysis of SDF-Net. The heatmaps illustrate the variation of mAP, on the left, and Rank-1, on the right in percentage (\%), under different combinations of the orthogonality constraint weight $\lambda_{\text{orth}}$ and the structure consistency weight $\lambda_{\text{struct}}$.}
    \label{fig:hyperparam}
\end{figure}

To thoroughly investigate the stability of the proposed framework, we conduct a sensitivity analysis on the two critical hyper-parameters: the orthogonality weight $\lambda_{\text{orth}}$ and the structural consistency weight $\lambda_{\text{struct}}$. As visualized in the heatmaps in Fig. \ref{fig:hyperparam}, the model performance exhibits robust tolerance across a wide range of parameter combinations.

The structural weight $\lambda_{\text{struct}}$ is optimal around 1.0, effectively bridging the modality gap without overwhelmingly dictating the semantic feature space. Concurrently, increasing the orthogonality weight $\lambda_{\text{orth}}$ progressively forces the disentanglement of shared and specific features, reaching peak discriminative capacity at $\lambda_{\text{orth}}=10.0$. Deviations from these optimal settings lead to gentle performance degradation rather than catastrophic failure, confirming that SDF-Net is not reliant on excessive hyper-parameter tuning and demonstrating strong generalization potential for cross-modal maritime retrieval tasks.

\subsection{Stability Analysis under Varying Physical Conditions}

To complement the main experiments, we evaluate SDF-Net's robustness under three physical parameters relevant to operational deployment: spatial resolution, SAR incidence angle, and target motion. All four ablation variants---baseline, SCL-only, DFL-only, and full SDF-Net---are compared under identical settings, isolating the contribution of each module to physical robustness.

\subsubsection{Robustness to Spatial Resolution}

The spatial resolution of satellite imagery varies significantly across different sensor systems and acquisition modes. While the HOSS-ReID dataset provides imagery at fixed native resolutions (0.75~m optical, 1.0~m SAR), operational scenarios frequently involve multi-resolution data sources. To evaluate SDF-Net's resilience to resolution mismatch, we synthesize degraded and enhanced versions of all test images using bilinear interpolation at scale factors of $\{0.25\times, 0.50\times, 0.75\times, 1.0\times, 1.50\times, 2.0\times\}$ relative to the native resolution. The model weights are frozen at the checkpoint trained on native-resolution data, ensuring that any observed robustness stems from architectural design rather than multi-resolution training.

\begin{table}[htbp]
    \centering
    \caption{Resolution stability with ablation comparison. SCL-only = baseline + SCL; DFL-only = baseline + DFL; SDF-Net = SCL + DFL. Bold entries mark the best value in each row. All metrics in percentage (\%).}
    \label{tab:stability_resolution}
    \renewcommand{\arrayrulewidth}{1pt}
    \setlength{\tabcolsep}{2.5pt}
    \begin{tabular}{ccccccccc}
        \toprule
        \multirow{2}{*}{Scale} & \multicolumn{2}{c}{Baseline} & \multicolumn{2}{c}{SCL-only} & \multicolumn{2}{c}{DFL-only} & \multicolumn{2}{c}{SDF-Net} \\
        \cmidrule(lr){2-3}\cmidrule(lr){4-5}\cmidrule(lr){6-7}\cmidrule(lr){8-9}
                               & mAP            & R1             & mAP            & R1             & mAP            & R1             & mAP            & R1 \\
        \midrule
        0.25$\times$           & 53.2           & 63.1           & \textbf{53.8}  & 61.9           & 52.4           & 62.5           & 53.1           & 62.5 \\
        0.50$\times$           & 56.0           & 64.2           & 57.3           & 64.2           & 57.1           & \textbf{67.0}  & \textbf{57.4}  & 65.9 \\
        0.75$\times$           & 57.5           & 64.8           & 57.3           & 64.2           & 58.8           & \textbf{68.8}  & \textbf{58.8}  & 67.0 \\
        1.00$\times$ (native)  & 58.6           & 67.6           & 59.2           & 66.5           & 59.8           & \textbf{69.9}  & \textbf{60.9}  & \textbf{69.9} \\
        1.50$\times$           & 57.8           & 66.5           & 58.0           & 64.8           & 58.9           & \textbf{68.2}  & \textbf{59.6}  & 67.6 \\
        2.00$\times$           & 57.8           & 66.5           & 58.0           & 64.8           & 58.9           & \textbf{68.2}  & \textbf{59.6}  & 67.6 \\
        \bottomrule
    \end{tabular}
\end{table}

As shown in Table~\ref{tab:stability_resolution}, SDF-Net achieves the best performance from native resolution down to $0.50\times$, while at $0.25\times$, or approximately 3.0~m GSD, SCL-only leads at 53.8\% versus 53.2\% for the baseline, consistent with SCL's role as the robustness module. DFL-only underperforms at this extreme because modality-specific features become unreliable when the ship spans only a few pixels. The upsampling experiments plateau because bilinear interpolation cannot recover sub-pixel detail absent from the native imagery.

\subsubsection{Incidence Angle and Projective Distortion}

SAR image formation is governed by the local incidence angle $\theta$, which controls both backscatter intensity and geometric distortion. The HOSS-ReID dataset was acquired under near-nadir LEO geometry, with an estimated $\theta < 25^\circ$, where layover and foreshortening are confined to tall superstructures and do not distort the hull footprint that SCL anchors upon. Within this envelope, SCL's global spatial averaging of gradient magnitudes (Eq.~2) is inherently tolerant to local pixel displacement from mild layover, and Instance Normalization (Eq.~3) decouples structural topology from incidence-angle-dependent backscatter intensity.

To extend robustness beyond near-nadir conditions, we propose a physics-informed projective augmentation for SAR images during training: range-axis scaling $s_y \sim \mathcal{U}[\sin 25^\circ, 1.0]$ simulates foreshortening, and range-axis shear $g_y \sim \mathcal{U}[0, h_{\max} / \tan 25^\circ]$ simulates layover for superstructure heights up to $h_{\max}\approx 30$~m. The parameters follow directly from SAR range-Doppler geometry~\cite{curlander1991synthetic}; under this training proposal, optical images would remain unmodified.

To quantify current architectural tolerance before augmentation training, we evaluate all four variants on test images uniformly warped at $\theta \in \{5^\circ, 15^\circ, 25^\circ, 35^\circ\}$. Since the warp is applied to both modalities, the reported results are a conservative estimate of robustness.

\begin{table}[htbp]
    \centering
    \caption{Robustness to SAR projective distortion---foreshortening and layover---with ablation comparison. $\theta$ denotes the incidence angle; shallower angles produce more severe distortion. All metrics in percentage (\%). Models are evaluated without geometric augmentation training.}
    \label{tab:stability_projective}
    \renewcommand{\arrayrulewidth}{1pt}
    \setlength{\tabcolsep}{2.5pt}
    \begin{tabular}{ccccccccc}
        \toprule
        \multirow{2}{*}{$\theta$} & \multicolumn{2}{c}{Baseline} & \multicolumn{2}{c}{SCL-only} & \multicolumn{2}{c}{DFL-only} & \multicolumn{2}{c}{SDF-Net} \\
        \cmidrule(lr){2-3}\cmidrule(lr){4-5}\cmidrule(lr){6-7}\cmidrule(lr){8-9}
                                  & mAP            & R1             & mAP            & R1             & mAP            & R1             & mAP            & R1 \\
        \midrule
        $5^\circ$                 & 11.4           & 15.3           & \textbf{13.3}  & \textbf{19.9}  & 13.1           & 18.8           & 12.2           & 18.2 \\
        $15^\circ$                & 31.5           & 43.8           & 31.2           & 42.0           & 31.4           & 44.9           & \textbf{32.6}  & \textbf{45.5} \\
        $25^\circ$                & 41.7           & 52.8           & 41.9           & \textbf{55.1}  & 42.3           & 54.5           & \textbf{42.6}  & \textbf{55.1} \\
        $35^\circ$                & 46.3           & 56.8           & 45.2           & 55.7           & 46.2           & 57.4           & \textbf{47.1}  & \textbf{58.5} \\
        \bottomrule
    \end{tabular}
\end{table}

Table~\ref{tab:stability_projective} shows that performance degrades monotonically as $\theta$ decreases. At $\theta=5^\circ$, the near-grazing case, SCL-only leads with 13.3\% mAP vs.\ 11.4\% baseline, consistent with the resolution and blur findings. At moderate angles, SDF-Net achieves the best results. The consistent SCL-only advantage across subsampling, low-pass filtering, and geometric shearing confirms that the structure consistency constraint is the robustness mechanism. These results also establish the baseline that projective augmentation during training must improve upon.

\subsubsection{Robustness to Target Motion}

Ship motion during SAR acquisition introduces azimuthal defocusing, or Doppler smearing, a modality-specific degradation with no optical counterpart. To quantify sensitivity to this effect, we apply horizontal motion blur kernels of increasing length $\{3, 5, 7, 9, 11\}$ pixels uniformly to all test images, simulating the range of Doppler-induced defocusing from slow-moving cargo vessels at 3--5 px, to fast patrol boats at 11 px. Since motion blur is applied to optical images as well---which do not suffer from azimuthal defocusing in reality---the reported robustness represents a conservative lower bound.

\begin{table}[htbp]
    \centering
    \caption{Robustness to simulated SAR azimuthal defocusing, or motion blur, with ablation comparison. All metrics in percentage (\%). SCL-only remains the top performer under strong blur.}
    \label{tab:stability_motion}
    \renewcommand{\arrayrulewidth}{1pt}
    \setlength{\tabcolsep}{2.5pt}
    \begin{tabular}{ccccccccc}
        \toprule
        \multirow{2}{*}{Blur (px)} & \multicolumn{2}{c}{Baseline} & \multicolumn{2}{c}{SCL-only} & \multicolumn{2}{c}{DFL-only} & \multicolumn{2}{c}{SDF-Net} \\
        \cmidrule(lr){2-3}\cmidrule(lr){4-5}\cmidrule(lr){6-7}\cmidrule(lr){8-9}
                                   & mAP            & R1             & mAP            & R1             & mAP            & R1             & mAP            & R1 \\
        \midrule
        0 (native)                 & 58.6           & 67.6           & 59.2           & 66.5           & 59.8           & \textbf{69.9}  & \textbf{60.9}  & \textbf{69.9} \\
        3                          & 58.1           & 66.5           & 59.0           & 65.9           & 58.8           & 67.6           & \textbf{59.7}  & \textbf{67.6} \\
        5                          & 57.6           & 65.3           & 57.8           & 64.8           & 57.5           & 66.5           & \textbf{58.4}  & \textbf{67.0} \\
        7                          & 57.4           & 66.5           & \textbf{57.5}  & 65.3           & 56.4           & \textbf{66.5}  & 57.0           & 65.3 \\
        9                          & 55.8           & 64.8           & \textbf{56.8}  & \textbf{65.9}  & 55.7           & \textbf{65.9}  & 55.9           & 65.3 \\
        11                         & 55.0           & 64.2           & \textbf{55.6}  & 63.6           & 55.0           & \textbf{66.5}  & \textbf{55.6}  & \textbf{66.5} \\
        \bottomrule
    \end{tabular}
\end{table}

Table~\ref{tab:stability_motion} shows the same pattern: SDF-Net leads under mild blur, while SCL-only dominates at 9--11 px, with 56.8\% versus 55.8\% for the baseline at 9~px. Azimuthal defocusing suppresses the fine-grained discriminative features that DFL relies on, while SCL's macroscopic hull edges survive moderate low-pass filtering. The overall degradation is modest (baseline drops from 58.6\% to 55.0\% mAP), indicating that the ViT backbone already provides substantial tolerance to localized defocusing. For high-speed vessel scenarios, SAR autofocus preprocessing~\cite{wahl2003sar} or the SCL-only configuration is recommended.

\subsection{Visualization and Qualitative Analysis}

To systematically demystify the internal representation mechanism and evaluate the practical efficacy of SDF-Net, we conduct a comprehensive multi-level qualitative analysis. We first visualize the spatial attention distribution to confirm that the network successfully anchors on modality-invariant structural priors while suppressing sensor-specific interference. Building upon this, we trace the evolutionary trajectory of intermediate feature maps to physically interpret the process of geometric abstraction and justify our architectural design choices. Finally, these theoretical insights are projected into the practical application domain by contrasting the ultimate retrieval results against the baseline, conclusively demonstrating the robustness of the proposed framework in bridging the complex optical--SAR modality gap.

\subsubsection{Class Activation Mapping Analysis}

\begin{figure}[htbp]
    \centering
    \includegraphics[width=\linewidth]{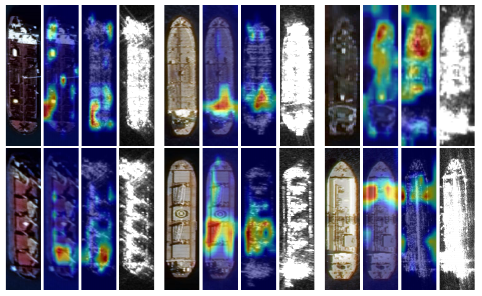}
    \caption{Grad-CAM visualization of the spatial attention maps generated by SDF-Net. From left to right within each group: the input optical image, the corresponding optical attention map, the SAR attention map, and the input SAR image. The network consistently focuses on the modality-invariant ship hull structure, effectively suppressing optical sea clutter and penetrating SAR speckle noise.}
    \label{fig:gradcam}
\end{figure}

To qualitatively validate the representation alignment mechanism of the proposed framework, we employ Grad-CAM to visualize the spatial attention distributions across heterogeneous sensing modalities. As illustrated in Fig. \ref{fig:gradcam}, SDF-Net yields highly concentrated and geometrically consistent activation responses across both the passive reflectance and active backscatter manifolds. Driven by the Structure-Aware Consistency learning module, the network consistently anchors its attention on the rigid hull contours and spatial layout of the maritime targets regardless of the imaging mechanism.

Specifically, within the optical domain, the network's attention successfully localizes the primary discriminative regions of the ships while effectively mitigating the influence of surrounding sea clutter. Crucially, the corresponding SAR activations exhibit a highly consistent spatial distribution. Despite the severe interference from inherent coherent speckle noise and high-intensity scattering artifacts, the model reliably attends to the corresponding spatial locations of the target. This strict cross-modal attention coherence substantiates the premise that enforcing intermediate structural constraints encourages the network to anchor its feature representations on modality-invariant spatial layouts rather than sensor-dependent textures, thereby establishing a robust identity association.

\subsubsection{Visual Analysis of Layer-wise Feature Evolution}

\begin{figure*}[htbp]
    \centering
    \includegraphics[width=0.5\textwidth]{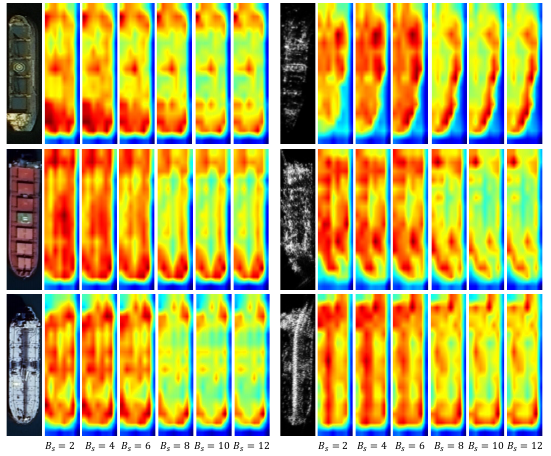}
    \caption{Visual evolution of feature heatmaps across different Transformer layers. For each modality group, the columns from left to right represent the original input image followed by the feature activation maps extracted from layers 2, 4, 6, 8, 10, and 12. Intermediate representations at layer 6 successfully isolate the modality-invariant geometric structure, whereas shallow layers are corrupted by sensor noise and deep layers suffer from spatial semantic collapse.}
    \label{fig:feature_evolution}
\end{figure*}

To elucidate the underlying mechanism of the Structure-Aware Consistency learning module, we examine the evolutionary trajectory of spatial representations by extracting feature heatmaps from varying depths of the Vision Transformer backbone. As presented in Fig. \ref{fig:feature_evolution}, the progression from shallow to deep layers reveals a distinct transition spanning from low-level sensor interference to high-level semantic abstraction.

In the early processing stages, specifically at layers 2 and 4, the feature maps exhibit high spatial resolution but remain severely entangled with modality-specific artifacts. Within the optical domain, the attention responses are heavily distracted by local background variations and illumination inconsistencies. Concurrently, the corresponding SAR features are dominated by discrete, high-intensity backscattering points and coherent speckle noise. This confirms that shallow token representations retain excessive sensor-dependent characteristics, rendering them unsuitable for direct geometric alignment.

As the hierarchical extraction progresses to the intermediate phase at layer 6, a profound spatial refinement occurs. The network effectively filters out the modality-specific radiometric distortions, distilling a concentrated and clean spatial layout of the ship. The physical proportions and topological structures are robustly preserved at this specific depth, providing an optimal and stable structural anchor for cross-modal matching. This visual phenomenon directly corroborates our prior quantitative ablation findings, which identify the sixth block as the optimal insertion point for the structural consistency constraint.

Conversely, the visualizations derived from the deeper stages spanning layers 8 through 12 demonstrate a progressive degradation of spatial fidelity. Driven by the terminal identity classification objective, the deep self-attention mechanisms aggressively aggregate global context. Consequently, the feature responses become spatially collapsed and overly abstract. Although these terminal representations encapsulate the highly discriminative semantic cues necessary for the final retrieval, they largely lose the fine-grained physical localization required for explicit structural regularization.

\subsubsection{Qualitative Retrieval Comparison}

\begin{figure*}[htbp]
    \centering
    \includegraphics[width=0.55\textwidth]{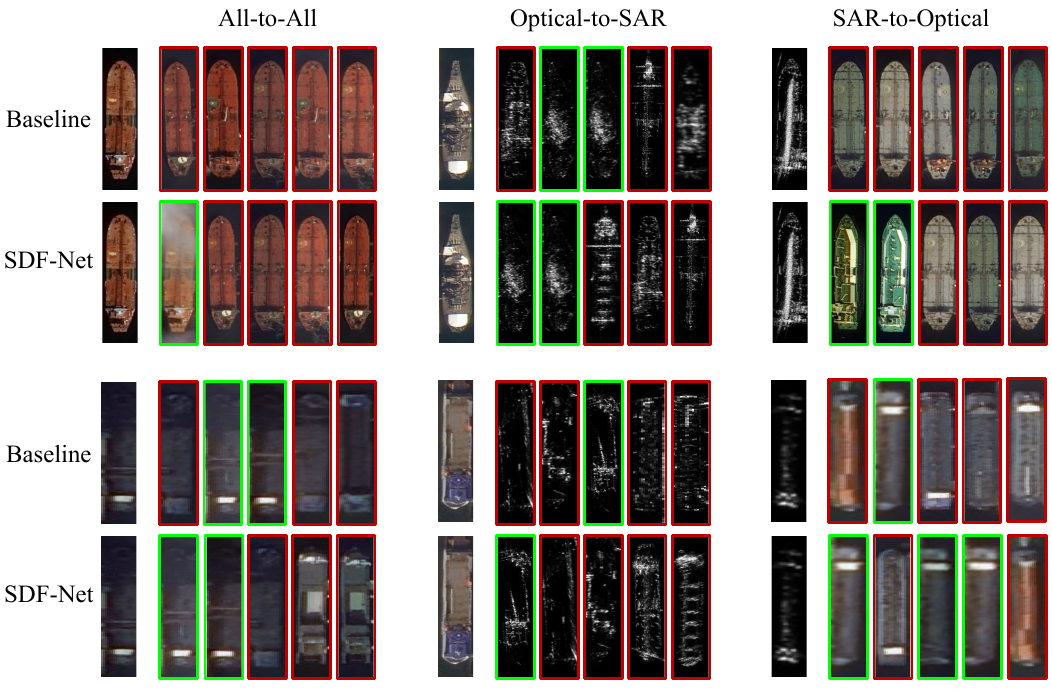}
    \caption{Qualitative comparison of retrieval results between the baseline and the proposed SDF-Net across three evaluation protocols. The left, middle, and right columns correspond to the All-to-All, Optical-to-SAR, and SAR-to-Optical settings, respectively. Green bounding boxes indicate correct matches, while red bounding boxes represent incorrect matches. SDF-Net exhibits superior robustness against modality-specific noise, consistently retrieving correct physical identities at higher ranks compared to the appearance-biased baseline.}
    \label{fig:retrieval_comparison}
\end{figure*}

To intuitively demonstrate the superiority of the proposed framework in bridging the heterogeneous sensing gap, we visualize the top-ranked retrieval results of SDF-Net alongside the baseline model. As illustrated in Fig. \ref{fig:retrieval_comparison}, the qualitative evaluation is conducted across the All, Optical-to-SAR, and SAR-to-Optical protocols.

The baseline model exhibits a pronounced vulnerability to the non-linear radiometric distortions inherent in cross-modal matching. Driven primarily by global appearance and texture similarities, the baseline frequently retrieves false positive candidates-indicated by red bounding boxes-at the highest ranks. This failure mechanism is particularly evident in the highly challenging Optical-to-SAR and SAR-to-Optical scenarios. Instead of matching the physical identity of the ships, the baseline tends to associate queries with gallery images sharing analogous maritime backgrounds, similar draft patterns, or comparable coherent speckle noise distributions. Such behavior underscores the fragility of purely statistical alignment when lacking explicit physical constraints.

Conversely, SDF-Net consistently identifies the correct targets, achieving a significantly higher density of true positive matches across all evaluated ranks and protocols. By anchoring the feature representation on modality-invariant geometric skeletons through the Structure-Aware Consistency learning module, the proposed network successfully circumvents the interference of complex sea clutter and radar artifacts. Even when querying with low-resolution SAR images highly corrupted by speckle noise, SDF-Net accurately retrieves the corresponding optical counterparts based on invariant structural primitives such as hull contours and spatial layouts.

\subsubsection{Failure Case Analysis}

\begin{figure}[htbp]
    \centering
    \includegraphics[width=0.6\linewidth]{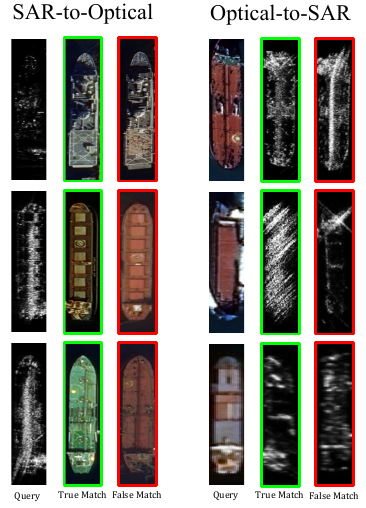}
    \caption{Failure case analysis of SDF-Net. Left panel: SAR-to-Optical retrieval. Right panel: Optical-to-SAR retrieval. Each row shows the query, the ground-truth match, and the incorrect top-1 prediction. Failures are predominantly caused by geometrically similar ship structures and low-resolution targets with weak structural cues.}
    \label{fig:failure_cases}
\end{figure}

To identify the boundary conditions of SDF-Net, we examine cases where the model fails to retrieve the correct identity at top rank. Fig.~\ref{fig:failure_cases} presents representative failures across both retrieval directions---SAR-to-Optical and Optical-to-SAR---with three cases per direction. Each row shows the query image, the correct ground-truth match, and SDF-Net's top-ranked but incorrect prediction.

Two recurring failure patterns emerge. The first involves ships with highly similar geometric profiles---comparable hull contours, aspect ratios, and superstructure layouts---that SDF-Net confuses because the SCL module anchors on macroscopic geometry rather than fine-grained texture. This is a direct consequence of the robustness--discriminability trade-off discussed in Section IV-E: SCL's structural anchoring suppresses modality-specific appearance cues that could otherwise disambiguate geometrically similar but distinct identities. The second pattern involves low-resolution targets where the ship spans too few pixels to generate reliable gradient energy statistics at the intermediate feature level, consistent with the 0.25$\times$ resolution boundary identified in the stability analysis. These failure modes define the practical limits of structure-centric cross-modal matching and motivate the future directions outlined in the Discussion.

\section{Discussion}

SDF-Net anchors optical--SAR ship ReID on modality-invariant geometric structures through two complementary modules: SCL extracts intermediate gradient energy to isolate the rigid hull from sensor-specific interference, while DFL factorizes terminal representations into shared and modality-specific subspaces fused via parameter-free additive refinement. On the HOSS-ReID benchmark, SDF-Net achieves 60.9\% mAP and 69.9\% Rank-1 accuracy under the All-to-All protocol, outperforming existing methods.

The ablation results point to a broader design principle: in active--passive cross-modal settings, disentanglement requires a stable geometric anchor to be beneficial. The stability analysis further suggests that intermediate gradient energy statistics---spatial derivatives, global averaging, and instance normalization---provide a robust and parameter-free structural descriptor that may generalize to other sensor pairs such as LiDAR--camera fusion.

Several limitations should be noted. The method degrades below approximately 3.0~m GSD and below $\theta \approx 15^\circ$ incidence, where geometric correction or projective augmentation training would be needed. DFL applied without SCL degrades performance; the modules must be deployed together. HOSS-ReID remains the only public benchmark in this domain, and generalization to additional ship types and ports awaits validation. Finally, SDF-Net assumes an upstream ship detector provides instance crops; detection errors will propagate to the ReID stage.

\section{Conclusion}

This work demonstrates that explicitly incorporating geometric structure as a physical prior into cross-modal representation learning---through intermediate gradient energy statistics and disentangled feature fusion---achieves state-of-the-art optical--SAR ship ReID while adding zero parameters. The complementary roles of SCL, which provides robustness under degradation, and DFL, which provides discriminative precision under clean conditions, establish a synergistic design paradigm for active--passive cross-modal matching. Future directions include integrating physics-informed projective augmentation into training, adaptive fusion weighting based on geometric anchor confidence, and extending the structure-centric paradigm to multi-view geometries beyond near-nadir observation.

\section{Acknowledgment}

The authors would like to thank all the researchers who kindly
shared the codes.

\bibliographystyle{IEEEtran}
\bibliography{reference}

@inproceedings{10687987,
    author    = {Xu, Qilong and Zhao, Xiuyang},
    booktitle = {2024 IEEE International Conference on Multimedia and Expo (ICME)},
    doi       = {10.1109/ICME57554.2024.10687987},
    number    = {},
    pages     = {1-6},
    title     = {Contour-Guided Modality Mitigation Network for Visible-Infrared Person Re-Identification},
    volume    = {},
    year      = {2024}
}

@article{8314449,
    author   = {Hughes, Lloyd H. and Schmitt, Michael and Mou, Lichao and Wang, Yuanyuan and Zhu, Xiao Xiang},
    doi      = {10.1109/LGRS.2018.2799232},
    journal  = {IEEE Geoscience and Remote Sensing Letters},
    number   = {5},
    pages    = {784-788},
    title    = {Identifying Corresponding Patches in SAR and Optical Images With a Pseudo-Siamese CNN},
    volume   = {15},
    year     = {2018}
}

@article{9416740,
    author   = {Fracastoro, Giulia and Magli, Enrico and Poggi, Giovanni and Scarpa, Giuseppe and Valsesia, Diego and Verdoliva, Luisa},
    doi      = {10.1109/MGRS.2021.3070956},
    journal  = {IEEE Geoscience and Remote Sensing Magazine},
    number   = {2},
    pages    = {29-51},
    title    = {Deep Learning Methods For Synthetic Aperture Radar Image Despeckling: An Overview Of Trends And Perspectives},
    volume   = {9},
    year     = {2021}
}

@article{9725265,
    author  = {Chen, Cuiqun and Ye, Mang and Qi, Meibin and Wu, Jingjing and Jiang, Jianguo and Lin, Chia-Wen},
    doi     = {10.1109/TIP.2022.3141868},
    journal = {IEEE Transactions on Image Processing},
    number  = {},
    pages   = {2352-2364},
    title   = {Structure-Aware Positional Transformer for Visible-Infrared Person Re-Identification},
    volume  = {31},
    year    = {2022}
}

@inproceedings{chen2023beyond,
    author    = {Chen, Weihua and Xu, Xianzhe and Jia, Jian and Luo, Hao and Wang, Yaohua and Wang, Fan and Jin, Rong and Sun, Xiuyu},
    booktitle = {Proceedings of the IEEE/CVF conference on computer vision and pattern recognition},
    pages     = {15050--15061},
    title     = {Beyond appearance: a semantic controllable self-supervised learning framework for human-centric visual tasks},
    year      = {2023}
}

@inproceedings{choi2020hi,
    author    = {Choi, Seokeon and Lee, Sumin and Kim, Youngeun and Kim, Taekyung and Kim, Changick},
    booktitle = {Proceedings of the IEEE/CVF conference on computer vision and pattern recognition},
    pages     = {10257--10266},
    title     = {Hi-CMD: Hierarchical cross-modality disentanglement for visible-infrared person re-identification},
    year      = {2020}
}

@article{dosovitskiy2020image,
    author  = {Dosovitskiy, Alexey},
    journal = {arXiv preprint arXiv:2010.11929},
    title   = {An image is worth 16x16 words: Transformers for image recognition at scale},
    year    = {2020}
}

@inproceedings{Fang_2023_ICCV,
    author    = {Fang, Xingye and Yang, Yang and Fu, Ying},
    booktitle = {Proceedings of the IEEE/CVF International Conference on Computer Vision (ICCV)},
    month     = {October},
    pages     = {11270-11279},
    title     = {Visible-Infrared Person Re-Identification via Semantic Alignment and Affinity Inference},
    year      = {2023}
}

@inproceedings{fu2021cm,
    author    = {Fu, Chaoyou and Hu, Yibo and Wu, Xiang and Shi, Hailin and Mei, Tao and He, Ran},
    booktitle = {Proceedings of the IEEE/CVF international conference on computer vision},
    pages     = {11823--11832},
    title     = {CM-NAS: Cross-modality neural architecture search for visible-infrared person re-identification},
    year      = {2021}
}

@inproceedings{gatys2016image,
    author    = {Gatys, Leon A and Ecker, Alexander S and Bethge, Matthias},
    booktitle = {Proceedings of the IEEE conference on computer vision and pattern recognition},
    pages     = {2414--2423},
    title     = {Image style transfer using convolutional neural networks},
    year      = {2016}
}

@inproceedings{he2021transreid,
    author    = {He, Shuting and Luo, Hao and Wang, Pichao and Wang, Fan and Li, Hao and Jiang, Wei},
    booktitle = {Proceedings of the IEEE/CVF international conference on computer vision},
    pages     = {15013--15022},
    title     = {Transreid: Transformer-based object re-identification},
    year      = {2021}
}

@article{hughes2018identifying,
    author    = {Hughes, Lloyd H and Schmitt, Michael and Mou, Lichao and Wang, Yuanyuan and Zhu, Xiao Xiang},
    journal   = {IEEE Geoscience and Remote Sensing Letters},
    number    = {5},
    pages     = {784--788},
    publisher = {IEEE},
    title     = {Identifying corresponding patches in SAR and optical images with a pseudo-siamese CNN},
    volume    = {15},
    year      = {2018}
}

@article{isprs-annals-III-1-9-2016,
    author  = {Ye, Y. and Shen, L.},
    doi     = {10.5194/isprs-annals-III-1-9-2016},
    journal = {ISPRS Annals of the Photogrammetry, Remote Sensing and Spatial Information Sciences},
    pages   = {9--16},
    title   = {HOPC: A NOVEL SIMILARITY METRIC BASED ON GEOMETRIC STRUCTURAL PROPERTIES FOR MULTI-MODAL REMOTE SENSING IMAGE MATCHING},
    volume  = {III-1},
    year    = {2016}
}

@article{li2022cross,
    author    = {Li, Liangzhi and Liu, Ming and Ma, Lingfei and Han, Ling},
    journal   = {International Journal of Applied Earth Observation and Geoinformation},
    pages     = {102964},
    publisher = {Elsevier},
    title     = {Cross-Modal feature description for remote sensing image matching},
    volume    = {112},
    year      = {2022}
}

@article{liang2024bridging,
    author    = {Liang, Tengfei and Jin, Yi and Liu, Wu and Wang, Tao and Feng, Songhe and Li, Yidong},
    journal   = {IEEE Transactions on Circuits and Systems for Video Technology},
    number    = {8},
    pages     = {7683--7698},
    publisher = {IEEE},
    title     = {Bridging the gap: Multi-level cross-modality joint alignment for visible-infrared person re-identification},
    volume    = {34},
    year      = {2024}
}

@inproceedings{Liu_2022_CVPR,
    author    = {Liu, Jialun and Sun, Yifan and Zhu, Feng and Pei, Hongbin and Yang, Yi and Li, Wenhui},
    booktitle = {Proceedings of the IEEE/CVF Conference on Computer Vision and Pattern Recognition (CVPR)},
    month     = {June},
    pages     = {19366-19375},
    title     = {Learning Memory-Augmented Unidirectional Metrics for Cross-Modality Person Re-Identification},
    year      = {2022}
}

@article{liu2020parameter,
    author    = {Liu, Haijun and Tan, Xiaoheng and Zhou, Xichuan},
    journal   = {IEEE Transactions on Multimedia},
    pages     = {4414--4425},
    publisher = {IEEE},
    title     = {Parameter sharing exploration and hetero-center triplet loss for visible-thermal person re-identification},
    volume    = {23},
    year      = {2020}
}

@inproceedings{liu2025advancing,
    author    = {Liu, Baolong and Huang, Roukai and Pan, Xin and Li, Chuanhuang and Sun, Jie and Dong, Jianfeng and Wang, Xun},
    booktitle = {Proceedings of the 48th International ACM SIGIR Conference on Research and Development in Information Retrieval},
    pages     = {106--115},
    title     = {Advancing Ship Re-Identification in the Wild: The ShipReID-2400 Benchmark Dataset and D2InterNet Baseline Method},
    year      = {2025}
}

@inproceedings{long2015learning,
    author       = {Long, Mingsheng and Cao, Yue and Wang, Jianmin and Jordan, Michael},
    booktitle    = {International conference on machine learning},
    organization = {PMLR},
    pages        = {97--105},
    title        = {Learning transferable features with deep adaptation networks},
    year         = {2015}
}

@article{merkle2018exploring,
    author    = {Merkle, Nina and Auer, Stefan and Mueller, Rupert and Reinartz, Peter},
    journal   = {IEEE Journal of Selected Topics in Applied Earth Observations and Remote Sensing},
    number    = {6},
    pages     = {1811--1820},
    publisher = {IEEE},
    title     = {Exploring the potential of conditional adversarial networks for optical and SAR image matching},
    volume    = {11},
    year      = {2018}
}

@article{pami21reidsurvey,
    author  = {Ye, Mang and Shen, Jianbing and Lin, Gaojie and Xiang, Tao and Shao, Ling and Hoi, Steven C. H.},
    journal = {IEEE Transactions on Pattern Analysis and Machine Intelligence},
    title   = {Deep Learning for Person Re-identification: A Survey and Outlook},
    year    = {2021}
}

@inproceedings{pan2018two,
    author    = {Pan, Xingang and Luo, Ping and Shi, Jianping and Tang, Xiaoou},
    booktitle = {Proceedings of the european conference on computer vision (ECCV)},
    pages     = {464--479},
    title     = {Two at once: Enhancing learning and generalization capacities via ibn-net},
    year      = {2018}
}

@inproceedings{park2021learning,
    author    = {Park, Hyunjong and Lee, Sanghoon and Lee, Junghyup and Ham, Bumsub},
    booktitle = {Proceedings of the IEEE/CVF international conference on computer vision},
    pages     = {12046--12055},
    title     = {Learning by aligning: Visible-infrared person re-identification using cross-modal correspondences},
    year      = {2021}
}

@inproceedings{Ren_2024_CVPR,
    author    = {Ren, Kaijie and Zhang, Lei},
    booktitle = {Proceedings of the IEEE/CVF Conference on Computer Vision and Pattern Recognition (CVPR)},
    month     = {June},
    pages     = {393-402},
    title     = {Implicit Discriminative Knowledge Learning for Visible-Infrared Person Re-Identification},
    year      = {2024}
}

@article{schmitt2019sen12ms,
    author  = {Schmitt, M. and Hughes, L. H. and Qiu, C. and Zhu, X. X.},
    doi     = {10.5194/isprs-annals-IV-2-W7-153-2019},
    journal = {ISPRS Annals of the Photogrammetry, Remote Sensing and Spatial Information Sciences},
    pages   = {153--160},
    title   = {SEN12MS – A CURATED DATASET OF GEOREFERENCED MULTI-SPECTRAL SENTINEL-1/2 IMAGERY FOR DEEP LEARNING AND DATA FUSION},
    volume  = {IV-2/W7},
    year    = {2019}
}

@inproceedings{sun2017svdnet,
    author    = {Sun, Yifan and Zheng, Liang and Deng, Weijian and Wang, Shengjin},
    booktitle = {Proceedings of the IEEE international conference on computer vision},
    pages     = {3800--3808},
    title     = {Svdnet for pedestrian retrieval},
    year      = {2017}
}

@inproceedings{touvron2021training,
    author       = {Touvron, Hugo and Cord, Matthieu and Douze, Matthijs and Massa, Francisco and Sablayrolles, Alexandre and J{\'e}gou, Herv{\'e}},
    booktitle    = {International conference on machine learning},
    organization = {PMLR},
    pages        = {10347--10357},
    title        = {Training data-efficient image transformers \& distillation through attention},
    year         = {2021}
}

@inproceedings{Wang_2025_ICCV,
    author    = {Wang, Han and Li, Shengyang and Yang, Jian and Liu, Yuxuan and Lv, Yixuan and Zhou, Zhuang},
    booktitle = {Proceedings of the IEEE/CVF International Conference on Computer Vision (ICCV)},
    month     = {October},
    pages     = {7873-7883},
    title     = {Cross-modal Ship Re-Identification via Optical and SAR Imagery: A Novel Dataset and Method},
    year      = {2025}
}

@article{xu2025cmshipreid,
    author    = {Xu, Congan and Gao, Long and Liu, Yu and Zhang, Qi and Su, Nan and Zhang, Shaoxuan and Li, Tianyu and Zheng, Xiaomei},
    journal   = {IEEE Journal of Selected Topics in Applied Earth Observations and Remote Sensing},
    publisher = {IEEE},
    title     = {CMShipReID: A Cross-Modality Ship Dataset for the Re-IDentification Task},
    year      = {2025}
}

@article{yang2022sar,
    author    = {Yang, Xi and Zhao, Jingyi and Wei, Ziyu and Wang, Nannan and Gao, Xinbo},
    journal   = {Pattern Recognition},
    pages     = {108208},
    publisher = {Elsevier},
    title     = {SAR-to-optical image translation based on improved CGAN},
    volume    = {121},
    year      = {2022}
}

@inproceedings{zhang2018unreasonable,
    author    = {Zhang, Richard and Isola, Phillip and Efros, Alexei A and Shechtman, Eli and Wang, Oliver},
    booktitle = {Proceedings of the IEEE conference on computer vision and pattern recognition},
    pages     = {586--595},
    title     = {The unreasonable effectiveness of deep features as a perceptual metric},
    year      = {2018}
}

@article{zhang2020ls,
    author    = {Zhang, Tianwen and Zhang, Xiaoling and Ke, Xiao and Zhan, Xu and Shi, Jun and Wei, Shunjun and Pan, Dece and Li, Jianwei and Su, Hao and Zhou, Yue and others},
    journal   = {Remote Sensing},
    number    = {18},
    pages     = {2997},
    publisher = {MDPI},
    title     = {LS-SSDD-v1. 0: A deep learning dataset dedicated to small ship detection from large-scale Sentinel-1 SAR images},
    volume    = {12},
    year      = {2020}
}

@article{zhang2021attend,
    author    = {Zhang, Shizhou and Yang, Yifei and Wang, Peng and Liang, Guoqiang and Zhang, Xiuwei and Zhang, Yanning},
    journal   = {IEEE Transactions on Image Processing},
    pages     = {8861--8872},
    publisher = {IEEE},
    title     = {Attend to the difference: Cross-modality person re-identification via contrastive correlation},
    volume    = {30},
    year      = {2021}
}

@inproceedings{zhang2023diverse,
    author    = {Zhang, Yukang and Wang, Hanzi},
    booktitle = {Proceedings of the IEEE/CVF conference on computer vision and pattern recognition},
    pages     = {2153--2162},
    title     = {Diverse embedding expansion network and low-light cross-modality benchmark for visible-infrared person re-identification},
    year      = {2023}
}

@article{zhang2025adaptive,
    author    = {Zhang, Yukang and Yan, Yan and Lu, Yang and Wang, Hanzi},
    journal   = {International Journal of Computer Vision},
    number    = {4},
    pages     = {2176--2196},
    publisher = {Springer},
    title     = {Adaptive middle modality alignment learning for visible-infrared person re-identification},
    volume    = {133},
    year      = {2025}
}

@inproceedings{zheng2019joint,
    author    = {Zheng, Zhedong and Yang, Xiaodong and Yu, Zhiding and Zheng, Liang and Yang, Yi and Kautz, Jan},
    booktitle = {proceedings of the IEEE/CVF conference on computer vision and pattern recognition},
    pages     = {2138--2147},
    title     = {Joint discriminative and generative learning for person re-identification},
    year      = {2019}
}

@article{zheng2022visible,
    author    = {Zheng, Huantao and Zhong, Xian and Huang, Wenxin and Jiang, Kui and Liu, Wenxuan and Wang, Zheng},
    journal   = {Electronics},
    number    = {3},
    pages     = {454},
    publisher = {MDPI},
    title     = {Visible-infrared person re-identification: A comprehensive survey and a new setting},
    volume    = {11},
    year      = {2022}
}

@article{zheng2024versatile,
    author    = {Zheng, Wei-Shi and Yan, Junkai and Peng, Yi-Xing},
    journal   = {IEEE Transactions on Pattern Analysis and Machine Intelligence},
    number    = {3},
    pages     = {1362--1380},
    publisher = {IEEE},
    title     = {A versatile framework for multi-scene person re-identification},
    volume    = {47},
    year      = {2024}
}

@inproceedings{zhu2017unpaired,
    author    = {Zhu, Jun-Yan and Park, Taesung and Isola, Phillip and Efros, Alexei A},
    booktitle = {Proceedings of the IEEE international conference on computer vision},
    pages     = {2223--2232},
    title     = {Unpaired image-to-image translation using cycle-consistent adversarial networks},
    year      = {2017}
}

@article{zhu2025hypergraph,
    author    = {Zhu, Jiacheng and Ge, Hongwei and Liu, Yuxuan and Wu, Chunguo and Fan, Jiulin},
    journal   = {Engineering Applications of Artificial Intelligence},
    pages     = {111286},
    publisher = {Elsevier},
    title     = {Hypergraph-driven soft semantics flexible learning for visible--infrared person re-identification},
    volume    = {158},
    year      = {2025}
}

@book{curlander1991synthetic,
    author    = {Curlander, John C. and McDonough, Robert N.},
    title     = {Synthetic Aperture Radar: Systems and Signal Processing},
    publisher = {John Wiley \& Sons},
    year      = {1991}
}

@article{wahl2003sar,
    author    = {Wahl, D. E. and Eichel, P. H. and Ghiglia, D. C. and Jakowatz, C. V.},
    title     = {Phase gradient autofocus---a robust tool for high resolution SAR phase correction},
    journal   = {IEEE Transactions on Aerospace and Electronic Systems},
    volume    = {30},
    number    = {3},
    pages     = {827--835},
    year      = {2003}
}
% \newpage

\section*{Biographies}

\begin{IEEEbiography}[{\includegraphics[width=1in,height=1.25in,clip,keepaspectratio]{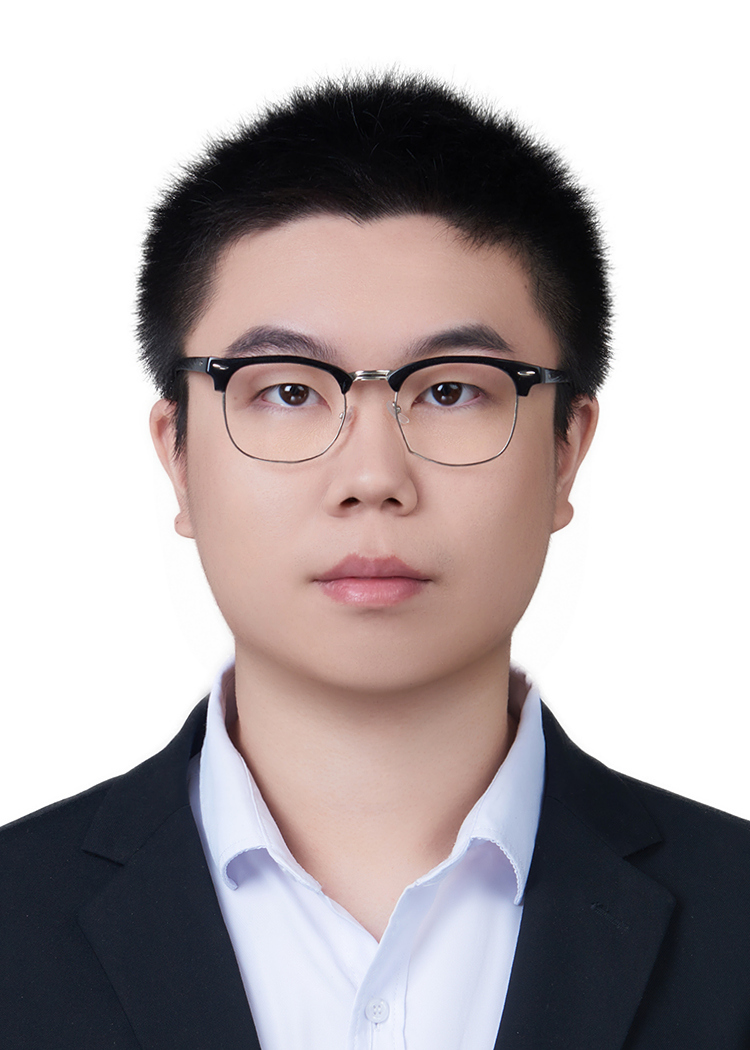}}]{Furui Chen}
received the B.M. degree in management science from Soochow University, Suzhou, China, in 2024. He is currently pursuing the M.S. degree in computer technology at the University of Chinese Academy of Sciences, Beijing, China.
His research interests include multimodal remote sensing analysis and deep learning.
\end{IEEEbiography}

\begin{IEEEbiography}[{\includegraphics[width=1in,height=1.25in,clip,keepaspectratio]{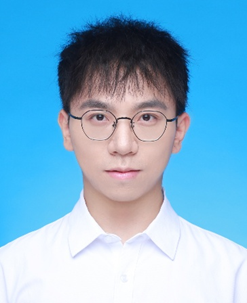}}]{Han Wang}
received the B.E. degree in electrical engineering from Chongqing University, Chongqing, China, in 2022. He is currently pursuing the Ph.D. degree in computer applied technology with the Technology and Engineering Center for Space Utilization, Chinese Academy of Sciences, Beijing, China.
His research interests include vision-language models, multimodal object detection and tracking, and remote sensing image interpretation.
\end{IEEEbiography}

\begin{IEEEbiography}[{\includegraphics[width=1in,height=1.25in,clip,keepaspectratio]{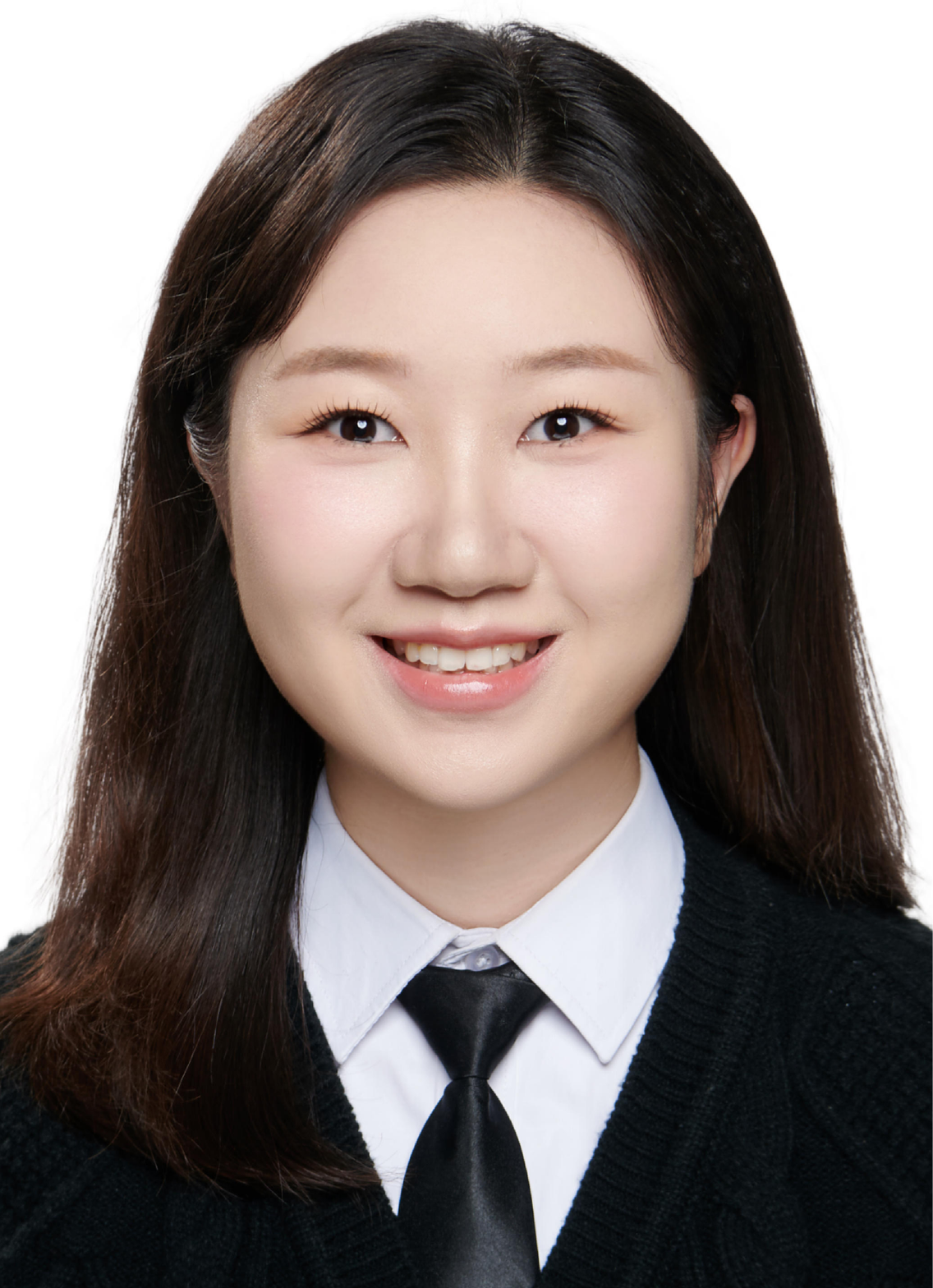}}]{Yuhan Sun}
received the B.S. degree in automation and computer science and technology from Xi'an Jiaotong University, Xi'an, China, in 2021. She is currently pursuing the Ph.D. degree in computer-applied technology with the Technology and Engineering Center for Space Utilization, Chinese Academy of Sciences, Beijing, China.
Her research interests include remote sensing image analysis, with a focus on object detection and vision-language model adaptation.
\end{IEEEbiography}

\begin{IEEEbiography}[{\includegraphics[width=1in,height=1.25in,clip,keepaspectratio]{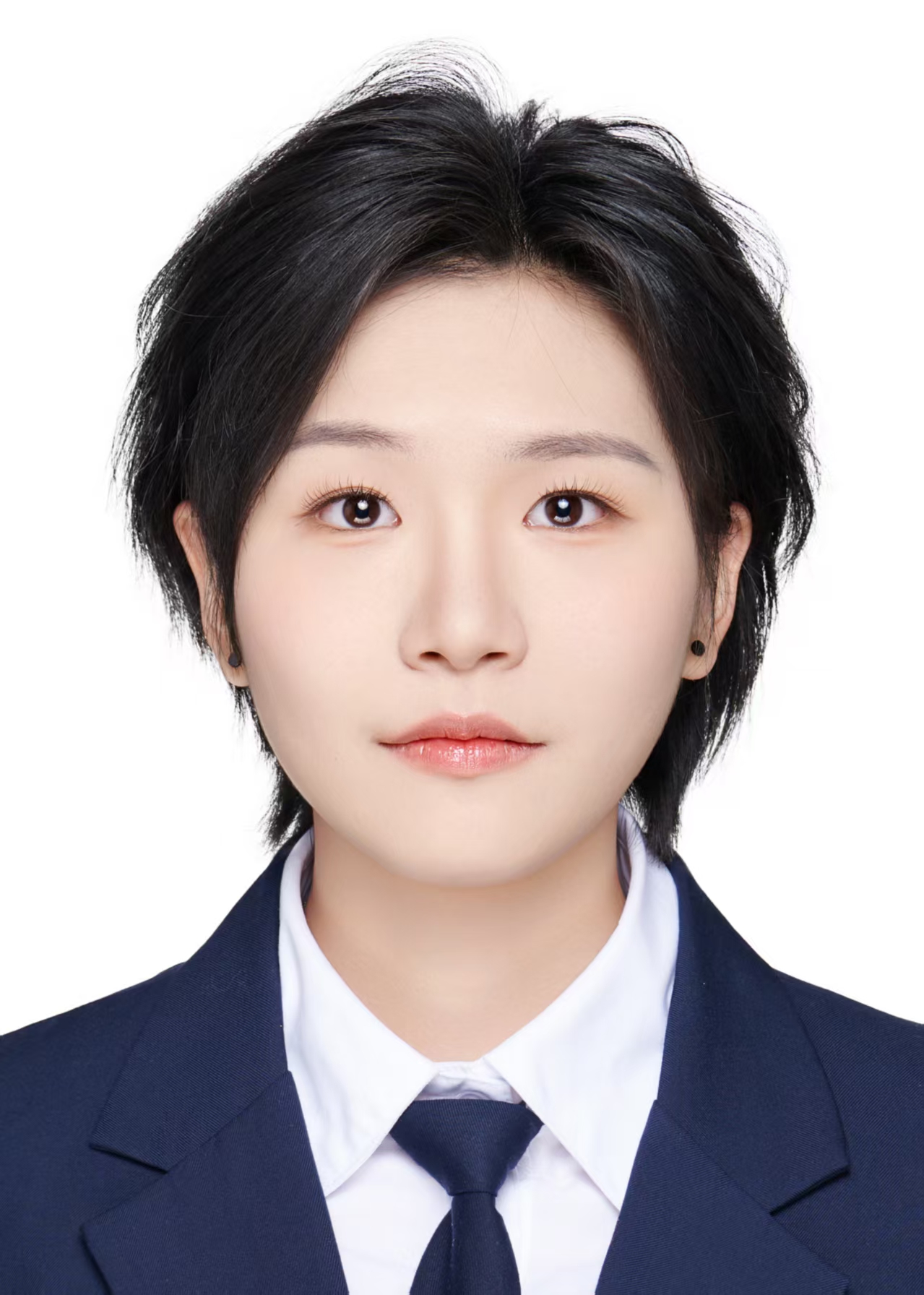}}]{Jianing You}
received the B.S. degree in artificial intelligence from China Agricultural University, Beijing, China, in 2024. She is currently pursuing the M.S. degree in computer technology at the University of Chinese Academy of Sciences, Beijing, China.
Her research interests include object detection, object tracking, and multimodal remote sensing image fusion.
\end{IEEEbiography}

\begin{IEEEbiography}[{\includegraphics[width=1in,height=1.25in,clip,keepaspectratio]{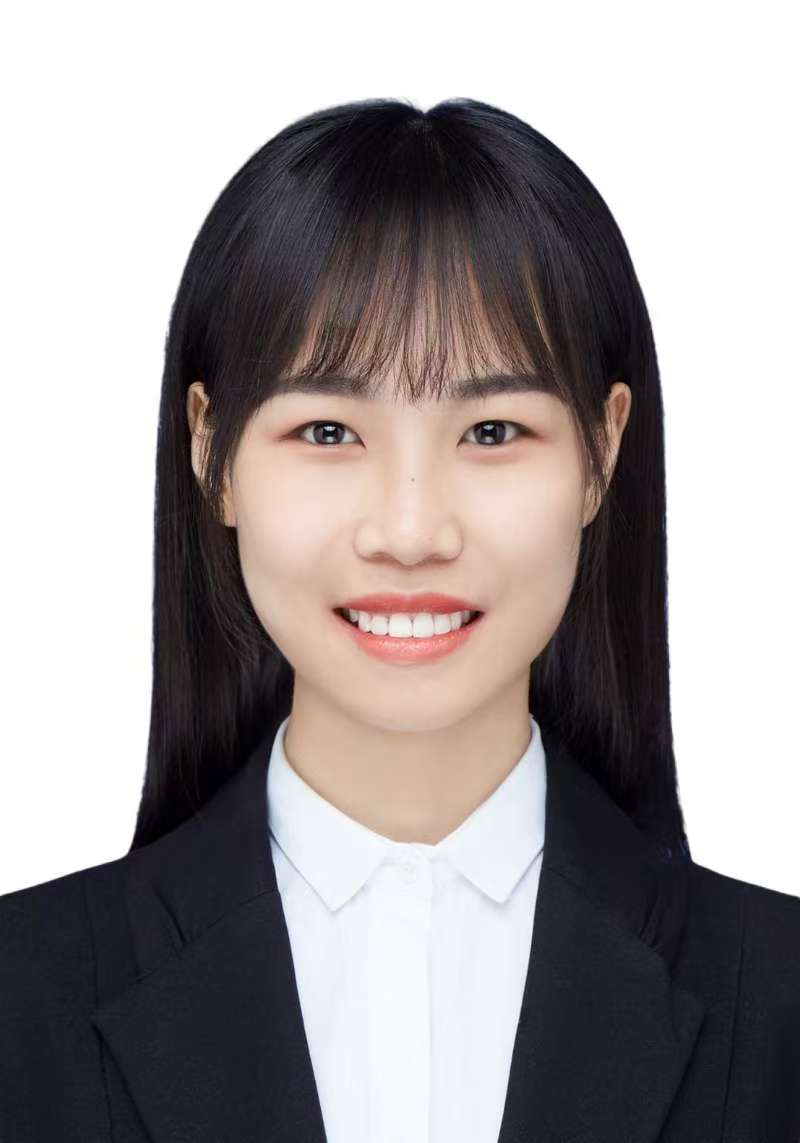}}]{Yixuan Lv}
received the B.Sc. degree from Xidian University, Xi'an, China, in 2019, and the M.Sc. degree in signal and information processing from the Aerospace Information Research Institute, Chinese Academy of Sciences, Beijing, China, in 2022. She is currently an Engineer with the Technology and Engineering Center for Space Utilization, Chinese Academy of Sciences, Beijing, China.
Her research interests include multimodal remote sensing analysis and deep learning.
\end{IEEEbiography}

\begin{IEEEbiography}[{\includegraphics[width=1in,height=1.25in,clip,keepaspectratio]{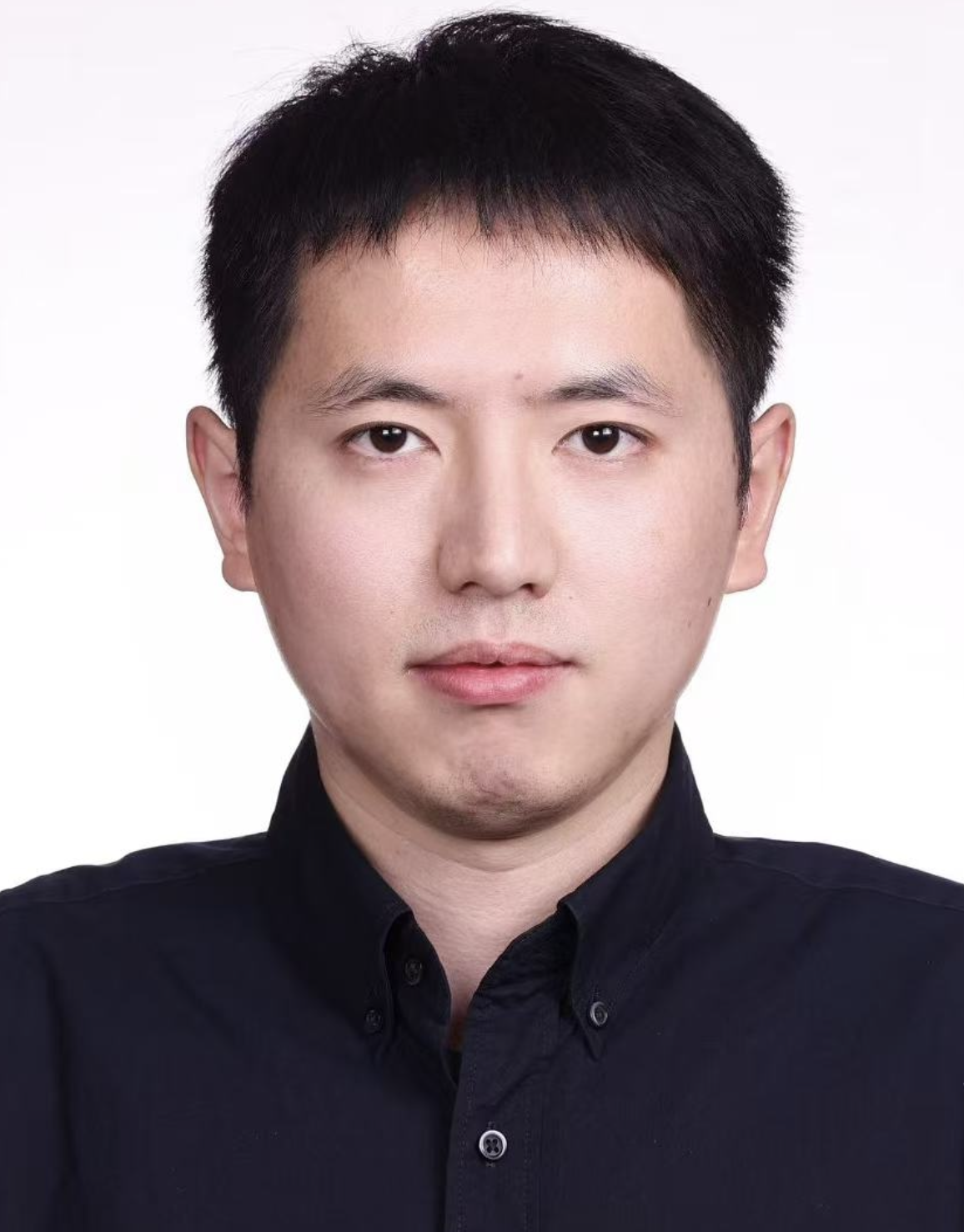}}]{Zhuang Zhou}
received the B.Eng. degree in electrical engineering and automation from the China University of Mining and Technology, Xuzhou, China, in 2013, the M.S. degree in cartography and geography information systems from Beijing Normal University, Beijing, China, in 2016, and the Ph.D. degree from the University of Chinese Academy of Sciences, Beijing, China, in 2024.
He is currently an Engineer with the Technology and Engineering Center for Space Utilization, Chinese Academy of Sciences, Beijing, China. His research interests include multi-source remote sensing fusion, specifically optical and SAR integrated processing, as well as intelligent image classification, detection, and tracking.
\end{IEEEbiography}

\begin{IEEEbiography}[{\includegraphics[width=1in,height=1.25in,clip,keepaspectratio]{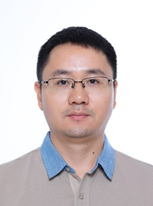}}]{Hong Tan}
received the M.S. and Ph.D. degrees in electronic engineering from the Institute of Electronics, Chinese Academy of Sciences, Beijing, China, in 2008 and 2016, respectively. Since 2016, he has been with the Key Laboratory of Space Utilization, Technology and Engineering Center for Space Utilization, Chinese Academy of Sciences, Beijing, China.
His research interests include data preprocessing, intelligent image processing, and data quality control for space applications.
\end{IEEEbiography}

\begin{IEEEbiography}[{\includegraphics[width=1in,height=1.25in,clip,keepaspectratio]{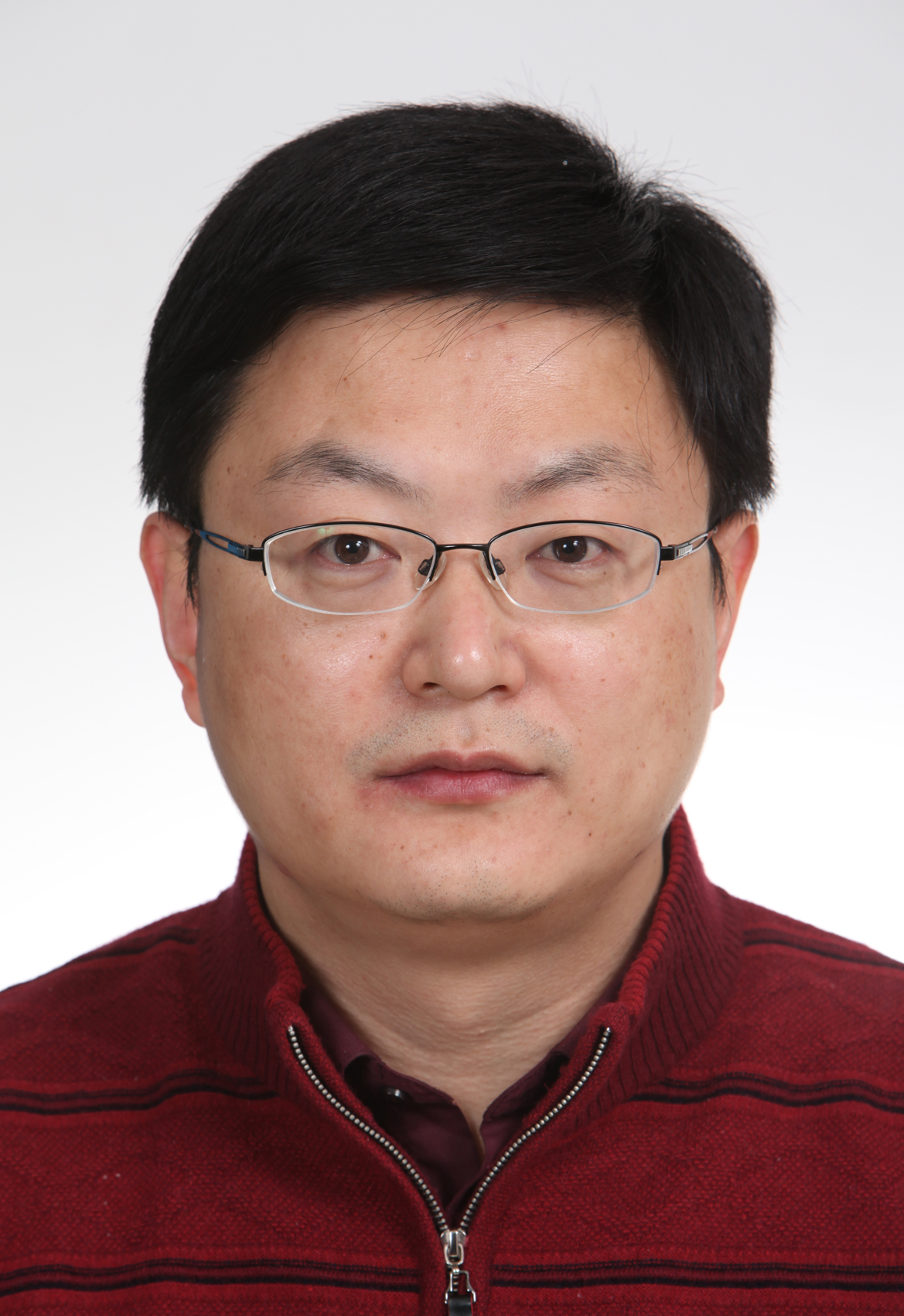}}]{Shengyang Li}
received the Ph.D. degree from the Institute of Remote Sensing Applications, Chinese Academy of Sciences, Beijing, China, in 2006. He is currently a Professor with the Technology and Engineering Center for Space Utilization, Chinese Academy of Sciences, Beijing, China.
His research interests include machine learning in remote sensing image interpretation, deep learning in satellite video processing and analysis, intelligent image processing for space utilization, and space scientific big data modeling and analysis.
\end{IEEEbiography}

\vfill

\end{document}